%% file: bare_jrnl_compsoc.tex
\documentclass[10pt,journal,compsoc]{IEEEtran}
\usepackage{cite}
   \usepackage[pdftex]{graphicx}
   \DeclareGraphicsExtensions{.pdf,.jpeg,.png}
\usepackage{tabularx, booktabs, ragged2e}

\usepackage{amsmath}
\usepackage[table,xcdraw]{xcolor}

\definecolor{diff}{rgb}{0.0, 0.0, 0.0}
\newcommand{\diff}[1]{\color{diff}{#1}~\color{black}}

\usepackage{algorithmic}

\usepackage{times}
\usepackage{comment}
\usepackage{soul}
\usepackage{tabularx}

\usepackage{soul}
\usepackage{url}
\usepackage[utf8]{inputenc}
\usepackage[small]{caption}
\usepackage{amsmath}
\usepackage{multirow}
\usepackage{adjustbox}
\usepackage{balance}
\usepackage[subrefformat=parens,labelformat=parens]{subcaption} 
\usepackage{balance}
\usepackage[]{algorithm2e}
\usepackage{tabularx, booktabs, ragged2e}
\usepackage{rotating}

\usepackage{enumitem}           
\usepackage{acronym}
\usepackage{ltablex, enumitem, makecell, float, textcomp}
\usepackage{array,colortbl}
\usepackage{bibunits}
\usepackage{comment}

\usepackage[pagebackref=true,breaklinks=true,letterpaper=true,colorlinks,bookmarks=false, citecolor=blue]{hyperref}

\usepackage{xspace}
\def\etc{etc.\@\xspace}
\newcommand{\etal}{\textit{et al}. }
\newcommand{\ie}{\textit{i}.\textit{e}., }
\newcommand{\eg}{\textit{e}.\textit{g}., }
\acrodef{gnn}[GNN]{graphical neural network}
\acrodef{fg}[FG]{IEEE International Conference on Automatic Face and Gesture Recognition}

\acrodef{ml}[ML]{machine learning}
\acrodef{sota}[SOTA]{state-of-the-art}

\acrodef{hog}[HOG]{histogram of gradients}
\acrodef{ss}[SS]{sister-sister}
\acrodef{bb}[BB]{brother-brother}
\acrodef{sibs}[SIBS]{brother-sister}

\acrodef{tfidf}[TF-IDF]{term frequency-inverse document frequency learning}

\acrodef{fs}[FS]{father-son}
\acrodef{ms}[MS]{mother-son}
\acrodef{fd}[FD]{father-daughter}

\acrodef{md}[MD]{mother-daughter}
\acrodef{gfgs}[GFGS]{grandfather-grandson}
\acrodef{gmgs}[GMGS]{grandmother-grandson}
\acrodef{gfgd}[GFGD]{grandfather-granddaughter}

\acrodef{gmgd}[GMGD]{grandmother-granddaughter}

\acrodef{sdm}[SDM]{signal detection model}
\acrodef{roc}[ROC]{receiver operating characteristic}
\acrodef{nmse}[NMSE]{Normalized Mean Square Error}
\acrodef{det}[DET]{Detection Error Trade-off}
\acrodef{tp}[TP]{true-positive}
\acrodef{tn}[TN]{true-negative}
\acrodef{ap}[AP]{average precision}
\acrodef{ae}[AE]{autoencoder}

\acrodef{bce}[BCE]{Binary Cross Entropy}
\acrodef{tpir}[TPIR]{true-positive identification rate}
\acrodef{frir}[FRIR]{false-reject identification rate}
\acrodef{fpir}[FRIR]{false-positive identification rate}

\acrodef{fn}[FN]{false-negative}
\acrodef{frr}[FRR]{false-reject rate}
\acrodef{fnr}[FNR]{false-negative rate}
\acrodef{fp}[FP]{false-positive}

\acrodef{fpr}[FPR]{\ac{fp} rate}
\acrodef{tpr}[TPR]{true-positive rate}

\acrodef{fiw}[FIW]{\textit{Families In the Wild}}
\acrodef{tsk}[TSKIN]{\textit{Tri-Subject Kinship}}

\acrodef{kfw}[KinFaceW]{\textit{Kin-Faces in the Wild}}
\acrodef{kfvw}[KFVW]{\textit{KinFaceW Videos}}

\acrodef{talkin}[TALKIN]{TALking KINship}

\acrodef{rfiw}[RFIW]{\textit{Recognizing Families In the Wild}}

\acrodef{nn}[NN]{neural network}

\acrodef{cnn}[CNN]{Convolutional Neural Network}
\acrodef{caae}[CAEE]{conditional adversarial \ac{ae}}
\acrodef{lut}[LUT]{Look-Up-Table}
\acrodef{fr}[FR]{face recognition}
\acrodef{gan}[GAN]{generative adversarial network}
\acrodef{dae}[DAE]{denoising \ac{ae}}

\acrodef{svm}[SVM]{Support Vector Machine}
\acrodef{mid}[MID]{Member ID}
\acrodef{fid}[FID]{Family ID}
\acrodef{pid}[PID]{Photo ID}
\acrodef{roc}[ROC]{receiver operating characteristic}
\acrodef{nrml}[NRML]{neighborhood repulsed metric learning}

\acrodef{fsp}[FSP]{`From same photograph'}
\definecolor{ao(english)}{rgb}{0.0, 0.5, 0.0}
\definecolor{review}{rgb}{0.0, 0.0, 0.0}
\definecolor{review2}{rgb}{0.0, 0.0, 0.0}

\definecolor{review3}{rgb}{0.0, 0.0, 0.0}

\acrodef{dkmr}[DKMR]{Deep Kinship Matching and Recognition}

\acrodef{lflkin}[LFL-KIN]{Linear Feature Learning}
\acrodef{hsl}[HSL]{Heterogeneous Similarity Learning}

\acrodef{sml}[SML]{Support Vector Data Description-based metric learning}
\acrodef{msml}[MSML]{multi-view SML}
\acrodef{lm3l}[LM$^3$L]{large-margin multi-metric learning}

\acrodef{svdd}[SVDD]{Support Vector Data Description}

\acrodef{tfidf}[TF-IDF]{term frequency-inverse document frequency}
\acrodef{nlp}[NLP]{Natural Language Processing}
\acrodef{map}[MAP]{mean average precision}

\usepackage{amssymb}

\begin{document}
%
\title{\color{review}Survey on the Analysis and Modeling of Visual \color{black}Kinship: A Decade in the Making}
%
%
%
%

\author{Joseph P Robinson,~\IEEEmembership{Student Member,~IEEE,}
        Ming Shao,~\IEEEmembership{Member,~IEEE,}
        and~Yun Fu,~\IEEEmembership{Fellow,~IEEE}
\IEEEcompsocitemizethanks{\IEEEcompsocthanksitem J.P. Robinson and Y. Fu were with the Department
of Electrical and Computer Engineering, Northeastern University, Boston,
MA, 02115.\protect\\
\IEEEcompsocthanksitem M. Shao was with University of Massachusetts, Dartmouth, MA, 02747.}
\thanks{Manuscript submitted as pre-print on February 24, 2021.}}

%
%

\markboth{Journal of PAMI, February~2021 (pre=print)}%
{Shell \MakeLowercase{\textit{et al.}}: Bare Demo of IEEEtran.cls for Computer Society Journals}
\IEEEtitleabstractindextext{%
\begin{abstract}
Kinship recognition is a challenging problem with many practical applications. With much progress and milestones having been reached after ten years - we are now able to survey the research and create new milestones. We review the public resources and data challenges that enabled and inspired many to hone-in on the views of automatic kinship recognition in the visual domain. The different tasks are described in technical terms and syntax consistent across the problem domain and the practical value of each discussed and measured. State-of-the-art methods for visual kinship recognition problems, whether to discriminate between or generate from, are examined. As part of such, we review systems proposed as part of a recent data challenge held in conjunction with the 2020 IEEE Conference on Automatic Face and Gesture Recognition. We establish a stronghold for the state of progress for the different problems in a consistent manner. This survey will serve as the central resource for the work of the next decade to build upon. For the tenth anniversary, the demo code is provided for the various kin-based tasks. Detecting relatives with visual recognition and classifying the relationship is an area with high potential for impact in research and practice.
\end{abstract}

}

\maketitle

\IEEEdisplaynontitleabstractindextext

\ifCLASSOPTIONpeerreview
\begin{center} \bfseries EDICS Category: 3-BBND \end{center}
\fi
%
\IEEEpeerreviewmaketitle

\IEEEraisesectionheading{\section{Introduction}\label{sec:introduction}}
\input{sections/1-introduction}

\input{sections/2-background}

\section{\color{review2}Visual Kinship Problems}\label{sec:data:benchmarks:resources}

\color{review}As mentioned, there are several kin-based tasks, each defined by specific protocols to best help control the experiment while simulating the use-case. Before we introduce the experimental details in the proceeding section, let us first introduce the various views of kin-based problems by the motivation and, thus, the problem statements for which they were found. Specifically, we review the discriminatory tasks (Fig.~\ref{fig:tasks}), along with the generative.
\color{black}

\subsection{Kinship verification}\label{subsec:verification}
The goal of the most popular kin-based task is to determine whether a face pair are blood relatives (\ie \emph{KIN} or \emph{NON-KIN}). Scholarly findings in the fields of psychology and computer vision revealed that different types of kinship share different familial features. From this, the task has evolved into verification over a broad range of relationship types, like \emph{parent}-\emph{child} (\ie \ac{fd}, \ac{fs}, \ac{md}, \ac{ms}) or siblings (\ie \ac{bb}, \ac{ss}, \ac{sibs}). Typically, we assume prior knowledge of the relationship type, both during training and testing. Hence, it is typical to train separate models or learn different metrics. Until the release of \ac{fiw}~\cite{robinson2016families, robinson2017recognizing}, small sample sets limited experiments. Thus, the larger data-pool of \ac{fiw} resulted in larger-scale evaluations that better mimicked true distributions of diverse families globally. With it, also came additional relationship types that span multiple generations (\ie \ac{gfgd}, \ac{gfgs}, \ac{gmgd}, \ac{gmgs}). \ac{fiw} is made-up of 1,000 disjoint family trees of various structures (\ie the number of family members range from five to forty-four). Furthermore, subject nodes making up the trees typically contain multiple face samples-- often samples that span over time, with face shots of most family members at different times in their lives. The families are split into five-folds with no overlap between folds.  The trees are converted to pairs of various relationship types, with an average of five face samples per family member. The pairs present a variety of additional challenges, as, for instance, a \ac{gmgs} pair may or may not be with faces of similar age. It could be an image of the grandmother as a child, and the GS as an adult, or even at the time he was a GF.

\color{review}Another flavor of kinship verification is best explained by the motivation behind UB Face: using knowledge of age as a prior and conditioned on whether or not \emph{KIN} is true~\cite{shao2011genealogical}. The idea was founded on analyzing the type of paired data frequently in the set of \ac{fp}. Specifically, facial pairs of relatives separated by larger age gaps. Thus, based on perceived hard positives, the UB Face dataset provided a pair of images per parent-- one at a younger and the other at an older age. In the end, Shao~\etal supported their hypothesis experimentally by showing a pair of true \emph{KIN} in \emph{parent}-\emph{child} relationships were closest when the parent was at a younger age. Then, Xia~\etal used this to formally claim \ac{sota} on UB Face by treating it as a transfer learning problem, with the target being that of the older parent and child, and the source being a younger version of the respective parent and child~\cite{Xia201144}. It is fairly known that paired data with greater age gaps are a challenge, and regardless of the level to which older children (\ie an elderly aged \emph{child}) compares to an older \emph{parent}.\color{black}

\begin{figure}[t!]
    \centering
    \begin{tabular}{c}
    \includegraphics[width=.9\linewidth]{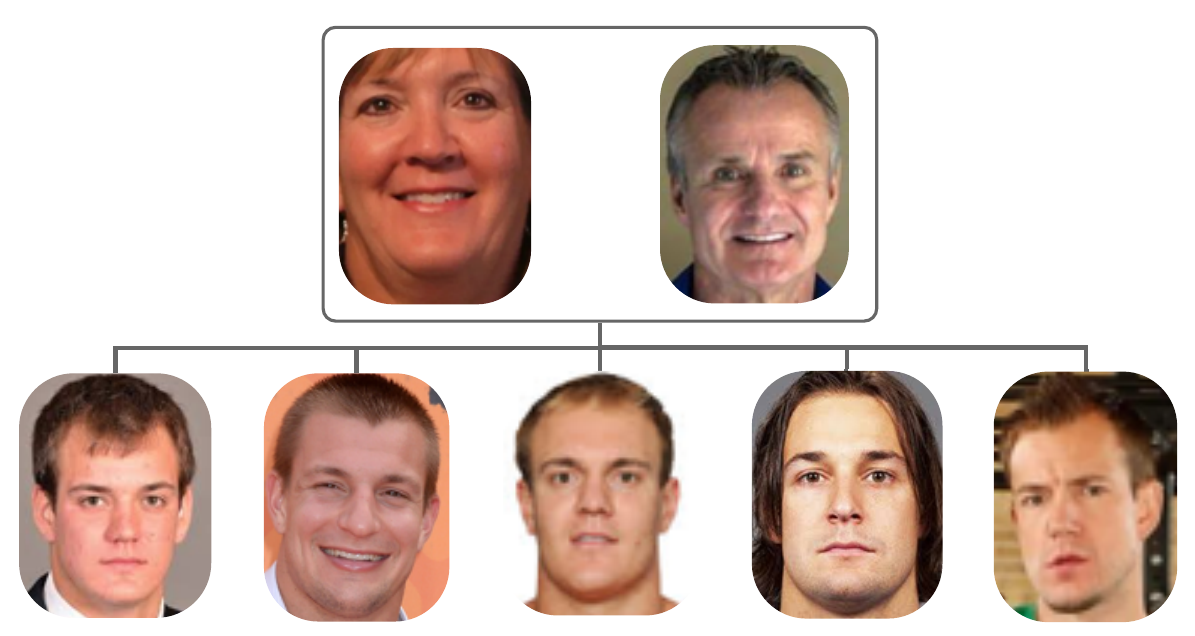}\\
    \includegraphics[trim=0 1mm 0 0, clip, width=.9\linewidth]{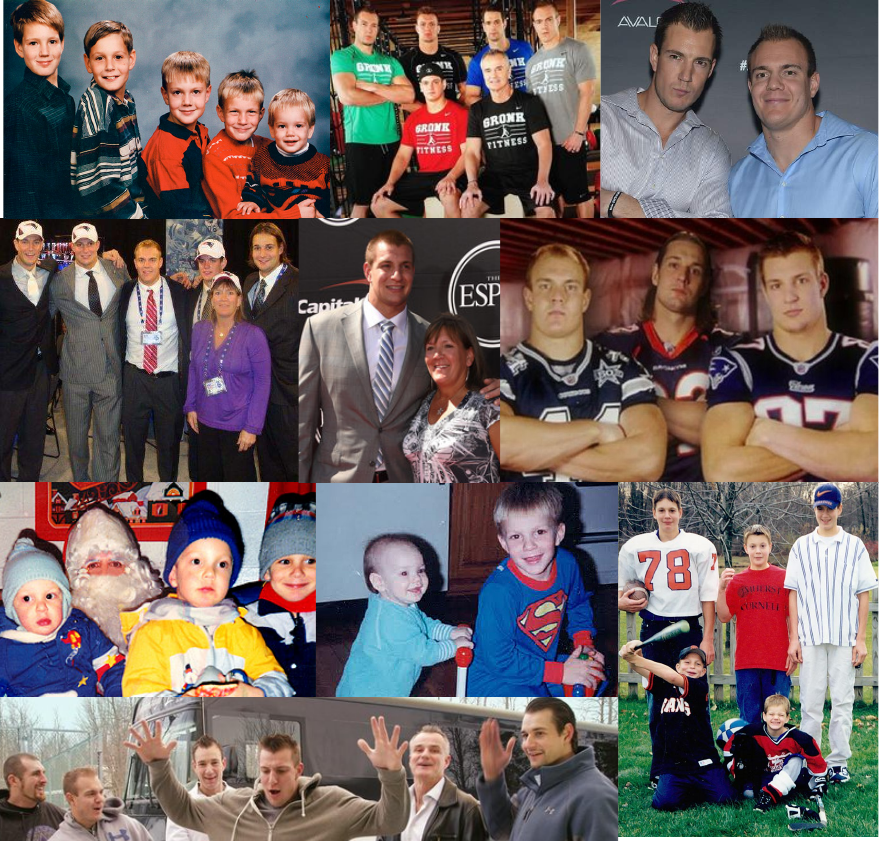}
     \end{tabular}
    \caption{\textbf{Sample family of \ac{fiw}~\cite{robinson2018visual}.} Faces and relationships of the American Football family, the Gronkowski's (\emph{Top}). The montage shows less than half of all photos for respective family. Photo types are various, spanning profile faces (top) to images of different subgroups of family members. Furthermore, samples capture different times of life. Note, crops were made to fit montage (\emph{Bottom}).}
    \label{fig:gronk:montage}
\end{figure}

\subsection{Family classification} Family classification, the problem where one family member is set aside, and all other members are used to model the classes (\ie family), is reviewed next. Hence, the task is to determine the family that the unknown subject belongs to, which is formulated as a closed-form, multi-class classification problem. This one-to-many problem is challenging, and only increases in difficulty with more families. The challenge stems from the large intra-class variations, which was revealed by a performance drop with an increasing number of families. Fang~\etal~\cite{fang2013kinship} was first to demonstrate this on Family101. Specifically, the authors showed a drop in performance from ten-to-fifty families (increments of 10); opposite to this, the performance improved with one-to-four (increment of one) family member during training. Robinson \etal included 316 families originally~\cite{robinson2016families}, then 512~\cite{robinson2018visual}, and finally 564~\cite{wu2018kinship}. After being supported as part of the \ac{rfiw} annual data challenge the first three consecutive years (\ie 2017, 18, and 19), the overview of the latest \ac{rfiw} mentioned the unrealistic setting of the problem, as families to employ on must be \emph{a priori} knowledge (\ie unable to generalize well). Thus, Robinson~\etal~\cite{robinson2020recognizing} omitted family classification from the latest challenge and substituted in two, more realistic views explained next. Nonetheless, as of the 2019 \ac{rfiw},  17.1\% (accuracy) is \ac{sota}~\cite{aspandi2019heatmap}.

\subsection{Tri-subject verification}
Tri-subject verification, introduced in~\cite{qin2015tri} (\ie \ac{tsk}), focuses on a slightly different view of verification. Specifically, the child is now paired with a parent pair, then asked whether \emph{KIN} or \emph{NON-KIN}. Hence, triplet pairs consist of a true parent pair with a child, \ie Father (F) / Mother (M) - Child (C) (FMC) pairs, where the child C is either Son (S) or Daughter (D) as FMS and FMD, respectively. TSKin makes the more practical sense, as knowledge of a single parent often means the other can be assumed. The last \ac{rfiw} used \ac{fiw} to support the largest benchmark~\cite{robinson2020recognizing}.

\subsection{Search and retrieval} 
This view, the most recent to be introduced~\cite{wu2018kinship}, formulates the problem of missing (\ie unknown) children. A search \emph{gallery} made up of all faces of \ac{fiw}, but those of the single child held out as the \emph{probes} for $F$ families. Thus, the input is visual media of an individual, and the output is a ranked list that includes all subjects in the \emph{gallery}. This \textit{many-to-many} task is staged as a \textit{closed set} problem. Thus, the number of \ac{tp} varies for each subject, ranging anywhere from $[1, k]$ relatives present in the \emph{gallery}. In other words, there are always relatives present.

\subsection{Multi-modal data}
Additional modalities (\eg video~\cite{kohli2018supervised, yan2018video}, audio~\cite{wu2019audio}\color{review}, multimedia~\cite{robinson2020familiesinMM}\color{black}), although limited attempts and fairly new in literature, have proven quite effective. \color{review}\ac{kfvw}, spawned out of the same group as \ac{kfw}, meaning notable contributions by these authors about half way through the decade (Fig.~\ref{fig:timeline}).  \color{black}Wu~\etal demonstrated that speech can be modeled to detect kinship~\cite{wu2019audio}. Audio, in particular, has shown promising, but through minimal attempts. To better understand the patterns that allow for speech to work-- whether that be jargon used, accents shared, or other acoustical features-- we have seen that a kinship detection system can be improved with audio; however, an in-depth look at the model and the salient components of highly matched signals is subject to future work.

\subsection{Kin-based facial synthesis}\label{subsec:synthesis}
Technology to post-process images  (or even curate in real-time, \ie Snapchat filters) have grown popular in the modern-day main-stream. 
From this alone-- kin-based face synthesis for entertainment and digital art is inherently employable. As a concrete example, Snap Inc. introduced the ability to predict the offspring from a pair of faces in the Snapchat app mid-2019. Surely, a natural curiosity. Furthermore, studies support links between DNA and appearance~\cite{walsh2016predicting}, meaning it possible.

Another, nearly default use-case for synthesizing faces based on kinship is in law enforcement to predict the appearance of an unknown perpetrator provided knowledge of kin. \color{review} Furthermore, missing family members (\eg a kidnapped child) with face images of years prior (\ie images of adolescence) could be used as prior knowledge, along with the appearances of family in their adulthood, to predict the face of that missing family member as an adult. \color{black}Also, nature-based studies where latent variables control the appearance of an offspring in a manner that allows for the analyzer to explore. And, projecting further in time, presumably, is its place in genetics. If genetics allows for, say, tweaking the fusion of male and female chromosomes to avoid deformities about the face of an offspring, the ability to visualize changes in appearance as a function of changes in latency, would likely be needed. 

\color{review}Some have attempted to predict the appearance of offspring in research (Section~\ref{sec:synthesis}, while others seek a way to commercialize (Section~\ref{sec:applications}): the former (\ie laboratory-style experimentation) and the latter (\ie applied in practical use-cases) are revisited in greater detail later in the paper.\color{black}

\input{sections/experimental}

\color{review}
\subsection{Generative modeling approaches}\label{sec:synthesis}

The dynamics of the offspring synthesis problem has a great distinction from tradition one-to-one mapping - two parents with directional relationships are input as prior knowledge to predict the appearance of their child. Such a two-to-one problem raises the question on how to best fuse knowledge from a pair of faces. Let us know even consider information for various family members - the fusion then should consider directed relationships inherent to family trees. Current face synthesizers conditioned on kinship assumes knowledge of one~\cite{ozkan2018kinshipgan} or both~\cite{gao2019will, ertugrul2017will} parents.

Ozkan~\etal proposed KIN-\ac{gan} to synthesize a child's face from a sample of a single parent~\cite{ozkan2018kinshipgan}. The problem is inherently difficult, for the variation embedded in many complex factors nearly changes from one sample-to-the-next. Nonetheless, trying to solve the problem with just one parent is insensible-- it takes two to tango in nature and, thus, such a formulation is out of scope before the problem is even started. 

\begin{figure}[t!]
\centering
\includegraphics[width=.95\linewidth]{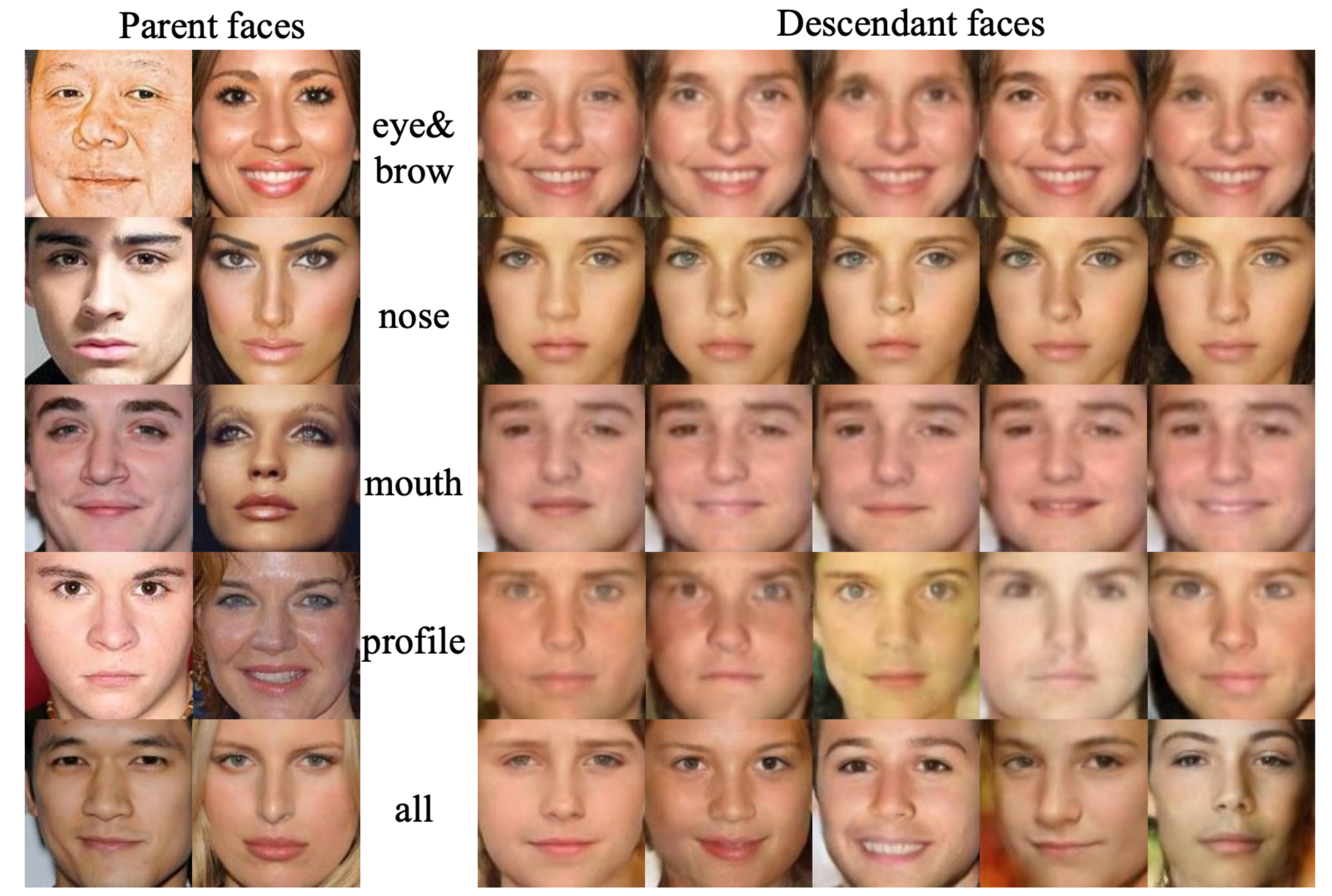}
\caption{\textbf{Child faces controlled by parts~\cite{zhang2020controllable}.} Best viewed in color.}
\label{fig:synthesizedparents}
\end{figure}

\begin{figure}[t!]
\centering
\begin{subfigure}[t]{0.35\linewidth}
\centering
\includegraphics[width=3cm]{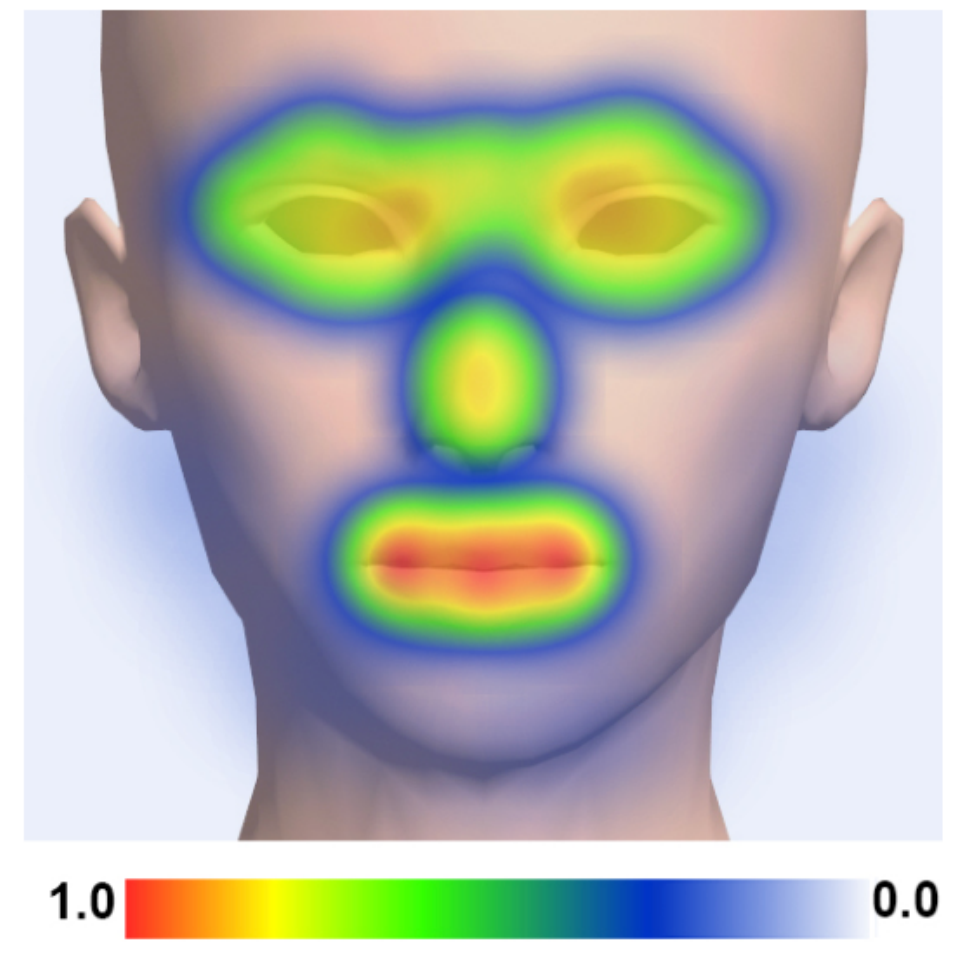}
\subcaption{Generated child}
\end{subfigure}
\quad
\begin{subfigure}[t]{0.35\linewidth}
\centering
\includegraphics[width=3cm]{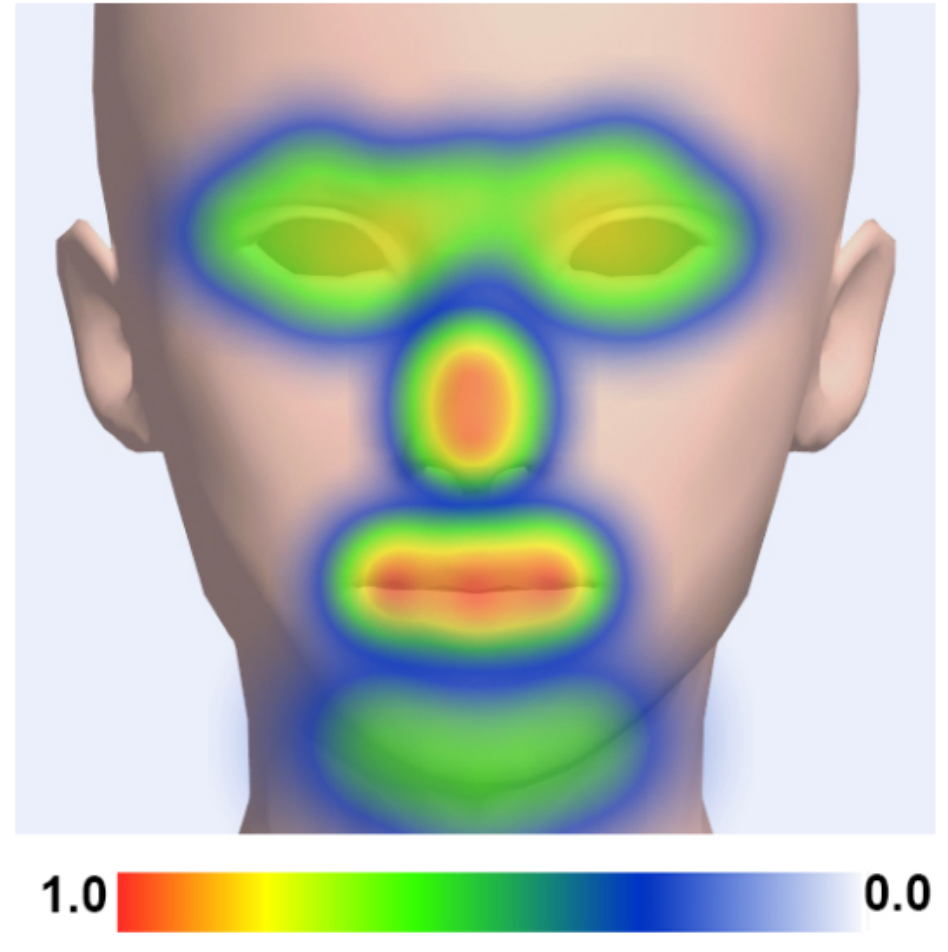}
\subcaption{Real child}
\end{subfigure}
\caption{\textbf{Salience map per key-points~\cite{gao2019will}.} Best viewed in color.}
\label{fig:heritablity}
\end{figure}

Ertugru~\etal proposed means of modeling as a two-to-one mapping~\cite{ertugrul2017will}. Similarly, Gao~\etal aimed to mimic the nature of reproduction with a model dubbed DNA-Net~\cite{gao2019will}. DNA-Net fuses latent representation of a parent pair at the feature-level, which is used as input to \ac{caae} model trained on top (Fig~\ref{fig:dna:net}). The parents signals are fused at the output of encoder E by concatenation of their features, and are then fed to the \ac{caae} model to produce a single feature representing the face encoding of the child. Finally, the child's encoding is decoded by G to the predicted facial image. Note that DNA-Net was dubbed by the authors in the effective work proposed; however fair when speaking in general terms (\ie infrequent situation in research), we suddenly see naming schemes such as this, \emph{genetic features}, among few others is too strong. Nonetheless, there is a clear analogy, so for the sake of story-telling and system depiction, Gao~\etal dubbed this as a single face is synthesized from face pair. The chose in \ac{caae} made it so the generator could synthesize children as a function of age and sex (Fig.~\ref{fig:dna:net:sample:results}). Note that treating sex as a continuous spectrum, opposed to discrete labels, is both appropriate and more precise (\ie provided an extreme pair, one female and the other male, there exists many cases in between, which is, in fact, where most of society falls~\cite{merler2019diversity}). GANKIN was another recent proposal that allowed for control of gender and age attributes~\cite{ghatas2020gankin}. Most recently, Zhang~\etal extended the ability of a model to control age and gender to one or more facial component for more diversity~\cite{zhang2020controllable} (Fig.~\ref{fig:synthesizedparents}). Returning to the work of DNA-Net, the authors compared salience in detecting kinship of type \emph{parent}-\emph{child} at specific facial features (\ie eyes, nose, mouth, and chin). Hu invariant moments were used as the shapes of the four facial parts~\cite{hu1962visual}, from which the accumulative cosine distances yielded \emph{heritability maps} (Fig.~\ref{fig:heritablity}). 
\color{black}

\section{Technical Challenges}\label{sec:challenges}
Like conventional \ac{fr}, unconstrained faces \emph{in the wild}~\cite{LFWTech} yield more difficult - imagery collected from sources outside a controlled laboratory environment is subject to more variations in pose, illumination, and scale. For faces, there are even more variables to further complicate the problem, such as expression and age. Furthermore, preparing to run such benchmarks to mimic real-world use-cases (\ie designing experiments and preparing the data) is, in itself, a challenge. Inheriting these challenges, but adding even more variations inherent in nature and in true data distributions of kinship, it is unsurprising that visual kinship recognition is a difficult problem. Nonetheless, great efforts over the last decade have been spent not just on solving the problems in kinship recognition, but also critiquing kinship research and its direction. We now elaborate on the challenges to keep this technology from making the transition of research-to-reality.

\color{review}
\subsection{Current limitations \ac{sota}}\label{sec:limitations}\color{black}
\color{review}
Still, we are close to achieving a performance-rating necessary for some applications (Section~\ref{sec:applications}). From this, we perceive that bridging the gap between research-and-reality (\ie transitioning from research-to-practice) is happening. Upon a clear assessment of the state of progress in research, we highlight barriers still in need of overcoming, along with sharing edge cases as means of highlighting common errors. Hence, we aim to inspire by explicitly depicting weaknesses in current \ac{sota} systems.

A clear limitation, however, is that most \color{black}solutions for visual kinship recognition assume the relationship type \emph{apriori}. Sometimes this could be practical, like if given a known source to decide whether or not the face, when paired with a target, is \emph{KIN} or \emph{NON-KIN}. Nonetheless, when considering the broader HCI incentive, along with data mining with social context, it is desirable to predict the exact type of relationship (\ie not just \emph{KIN} or \emph{NON-KIN}). Nonetheless, a high confidence in knowing whether a relationship does exist could serve as powerful prior knowledge when classifying the specific type.

\color{review}
Let us now consider the renowned \ac{kfw} dataset. Although the dataset has had a great impact in research, for having attracted many to the problem and, thus, has motivated many outstanding works, there are a few clear flaws in relating the results to real-world data. More than half of all true pairs making up \ac{kfw} are faces from the same photo. Researchers have then questioned the validity of the patterns being learned, showing that naive approaches such as color features~\cite{lopez2016comments} or detecting whether or not faces are from the same photo~\cite{dawson2018same} outperform \ac{sota} on most datasets, including \ac{kfw}. Thus, another clear limitation of some data resources is in the data distribution itself -- a technical challenge we soon cover in-depth (Section~\ref{subsec:datachallenges}).
\color{black}



\subsection{The nature}

\color{review}
\noindent\textbf{Demographics and inherent bias.}
\color{review}A challenge are issues of bias in \ac{fr} machinery. However, no study of bias in demographics (\eg ethnicity and gender) for kin-based data: a study that should be conducted, like Robinson~\etal did in finding variations in score sensitivities across subgroups in \ac{fr}~\cite{robinson2020face}.

\begin{figure}[!t]
    \centering
    \includegraphics[width=.9\linewidth]{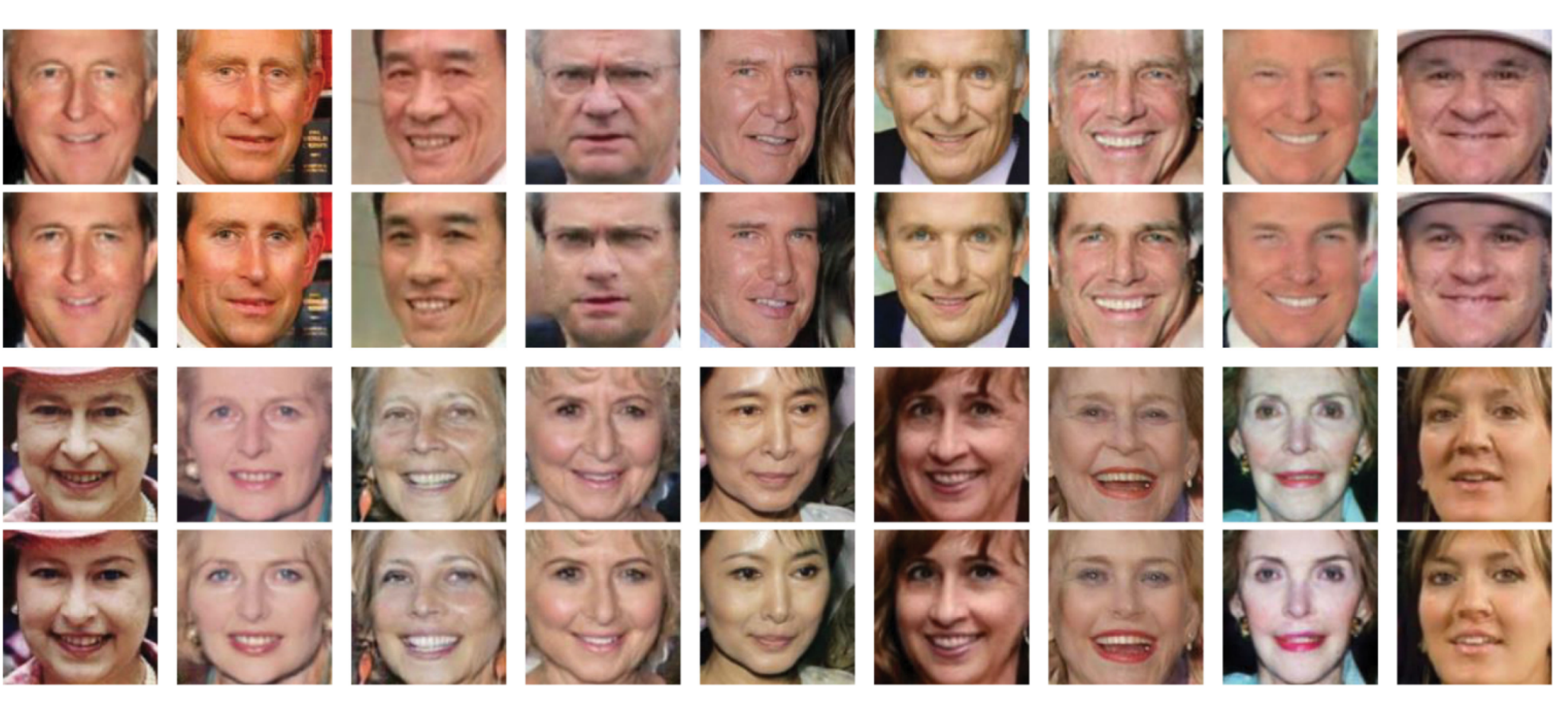}
    \caption{\textbf{Sample faces synthesized to improve predictive power for faces of elderly adults (visualization from~\cite{wang2018cross}).} Two models (\ie one per gender) were trained to synthesize input faces as younger-- male fathers (\ie rows 1-2) and female mothers (\ie rows 3-4), the top sample is the original and the generated is below.}
    \label{fig:cross:generation}
\end{figure}

\noindent\textbf{Effects of age variations.}
Family members with a large age-gap makes for more of a challenge. Wang~\etal demonstrated a benefit in having a face image synthesized at younger ages~\cite{wang2018cross}. Their ablation study revealed cumulative improvements as $x\sim p_Y$ was bounded to $>$20 years of age, then to $>$30, and up to $>$50. Improved results came with increasing the size of the domain (\ie the respective age considered young, which is orthogonal to those considered old). Fig.~\ref{fig:cross:generation} depicts samples of parents synthesized for kinship verification. Other data augmentation techniques have also proved useful, like transforming faces to their basis to then invert, rotate, and change ocular geometry~\cite{dal2015allocentric}.
\color{black}

\subsection{The environment}
The challenges from age variations in \ac{fr} not only intensify in kin-based problems, but also change in novel ways. For instance, let us assume a comparison in the faces of a grandmother and a prospective grandson. The age of each and age gap between the two are subject to variation. In other words, the problem inherits the same challenges of \ac{fr} such that considerations for directed relationships of concern-- the grandmother might be in her early years when the picture was captured, just as the grandson might even be a grandfather himself at the time the picture was taken.

Nurture adds additional challenges to the problem: For instance, a pair of brothers inherited the nose from their mother; one boy experienced a broken nose perhaps more than once; suddenly, that boy no longer has a nose that resembles the mother. Where such challenges exist in conventional \ac{fr}, the relative cost is greater with losing an inherited distinguishable feature from a prospective parent(s) in kin-based problems.

Biology-based research has focused on the problem of kinship recognition from a vast array of viewpoints. For instance, work that precedes the work done in machine vision, focused on a human's ability to recognize kinship-- specifically, the ability of younger siblings to better distinguish between \emph{KIN} and \emph{NON-KIN} in strangers\cite{kaminski2010firstborns} (\ie having seen the first-born their entire lives trains them). An interesting hypothesis indeed, which is supported in the reported experiments (minimal sample set, but typical of human evaluations done in face-based research). Intuitively, the contrary could also be true (\ie the role of the older sibling, watching after their younger sibling would better train for this ability). In any case, the authors propose a theory conditioned on age; difference in age could play a significant role in such a study, as we agree this could be the case for a much older sibling (\ie already developing an ability to discriminate between faces), the same argument of realizing the key differences as a means of recognizing kin in a sibling at a young age could be argued both ways. Furthermore, the authors discarded samples of subjects with no siblings and more than two siblings-- on the one hand the intent to control the experiment with less variation is understandable - on the other, subjects without siblings would serve as a meaningful baseline, while those with a number of siblings only strengthens the case for the oldest being the most keen on recognizing kin (\ie having grown watching over their younger siblings).

\subsection{The data and its distribution}\label{subsec:datachallenges}
Within-family variations are vast. As such, one cannot infer that the inherited traits from one father-son pair would mimic inherited traits of another father-son pair. Furthermore, the factors introducing added complexity vary across different ethnic groups. 

\begin{figure}
    \centering
    \includegraphics[width=\linewidth]{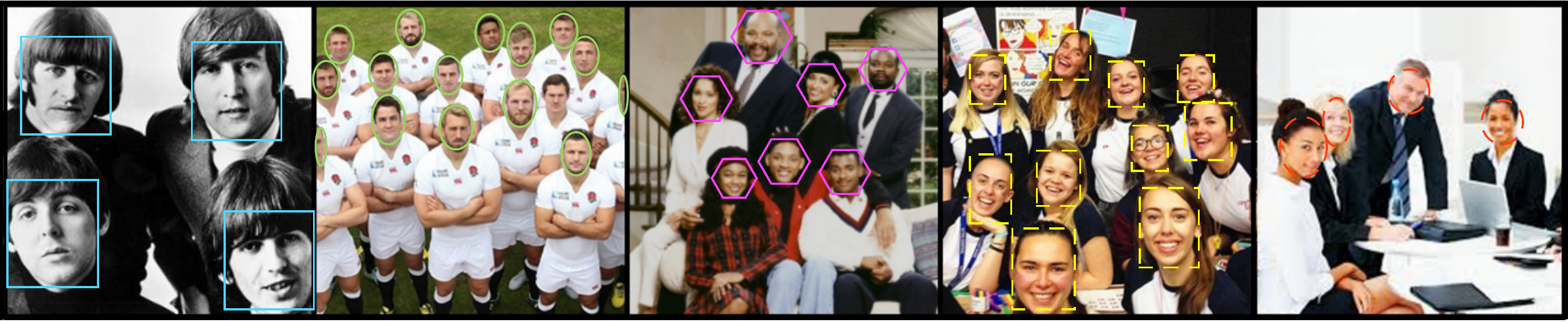}
    \caption{\textbf{\ac{fsp} data (modified from~\cite{dawson2018same}}). Tagged using \emph{must} and \emph{cannot}-link constraints, FSP had $\approx$1M samples from the web via 125 non-kin searches like \emph{school student}, \emph{business meeting}, \emph{team photos}.}
    \label{fig:fspdata}
\end{figure}

To capture the true data distributions of visual kinship as seen around the world is a great challenge, where many efforts have exhibited exploitable flaws. For instance, using color features claimed \ac{sota} on the \ac{kfw} dataset, as faces of true-relatives often were cropped from the same photos~\cite{lopez2016comments}\cite{wu2016usefulness}. The same motivation ushered in a different paradigm as means to measure unintended data leakage in the unnatural domain inherited by samples being of the same image or different. To say the least - this was a crafty piece of work that acquired an abundance of cheap data by image-level constraints that impose faces in the same photo as \emph{matches}, which means it is a binary problem with classes for the \emph{same} and \emph{different} photo. In other words, by the paired data acquired by finding images with one-to-many faces from the web (Fig.~\ref{fig:fspdata}), Dawson~\etal proposed training a detector to determine whether a face pair was from the \emph{same} or \emph{different} photo. Then, the boolean class model was directly evaluated on kin-based image sets, with the only difference in the target classes (\ie \emph{same} and \emph{different} assumed to be \emph{KIN} and \emph{NON-KIN}). Thus, showing \ac{sota} ratings on a majority of existing kinship data-- again, hypothesis that public benchmarks were subject to unintended data leakage, and one that is intrinsic to the distribution of classes (\ie \emph{KIN} and \emph{NON-KIN}). In the end, \ac{fsp} proved competitive on KFW-I, KFW-II, Cornell KF, and \ac{tsk}; however, \ac{fsp} lacks sufficient training to perform well on the multi-image \ac{fiw} data (\ie 58.6\%, which was the first, smallest version of the \ac{fiw} dataset). In fact, at the core of \ac{fiw} specifications, as defined in its earliest paper~\cite{robinson2016families}, the concept of same and different photo was one considered in the creation of \ac{fiw}-- mentioned as part of motivation for the data in other recent literature reviews on kin-based image datasets~\cite{dawson2018same}.

\input{sections/applications}

\input{sections/discussion}

%
\bibliographystyle{IEEEtran}
\bibliography{IEEEabrv,references}




%

\vspace{-10mm}
\begin{IEEEbiography}[{\includegraphics[width=1in,height=1.2in,clip,keepaspectratio]{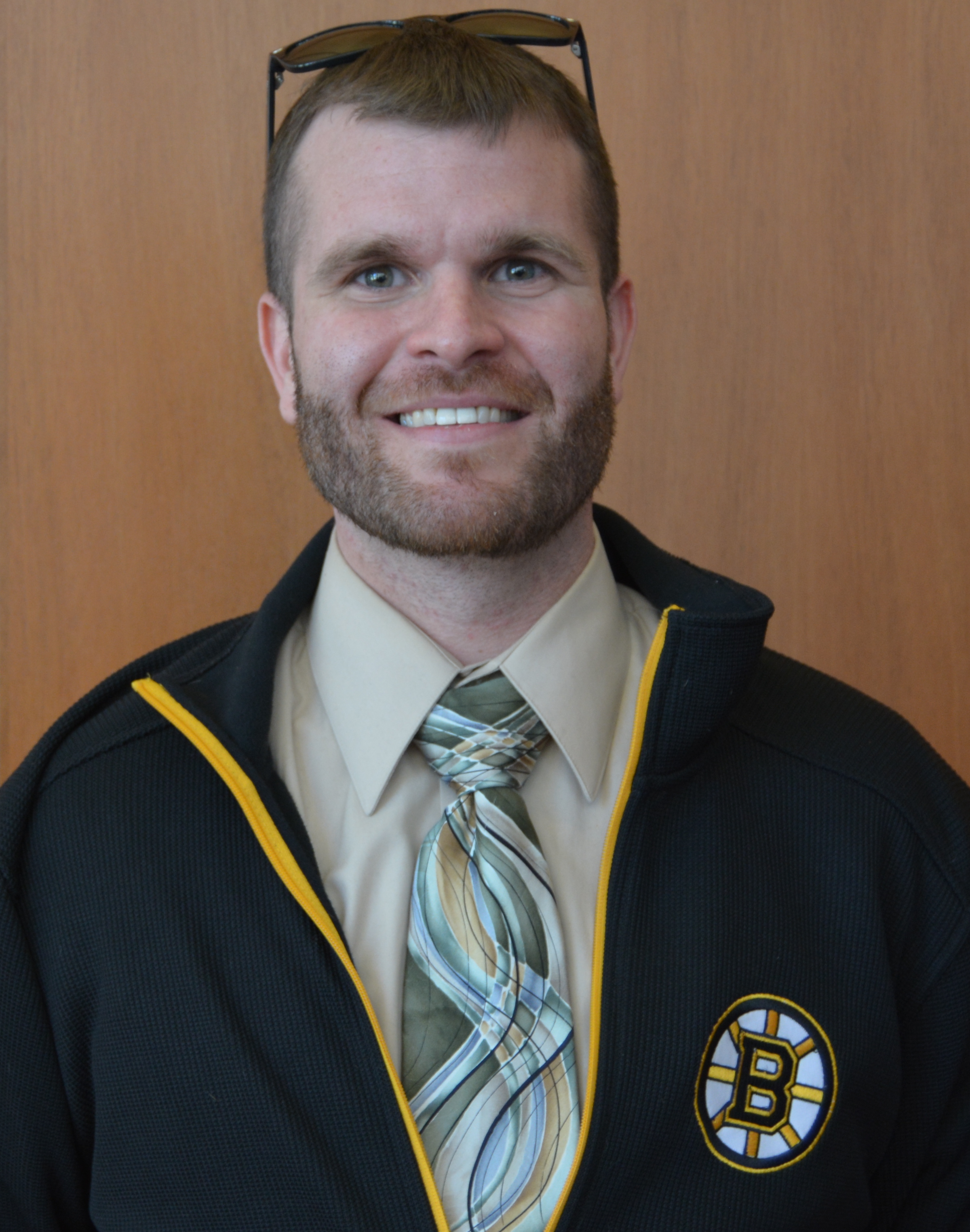}}]{Joseph P Robinson}
B.S. in electrical \& computer engineering ('14) and Ph.D. in computer engineering ('20) and part-time faculty ('17-'20) at Northeastern University-- designed and taught undergrad Data Analytics ('20~\emph{Best Teacher}). Research in applied machine vision, with emphasis on faces, deep learning, MM, big data. Led on TRECVid debut (MED'15, 3rd place). Built many image and video datasets-- most notably FIW. Served as Virtual Chair of FG, SPC of IJCAI; organized \& hosted several workshops and challenges (\eg NECV17, RFIW@ACMMM17, RFIW@FG18-20, AMFG@CVPR18-21, FacesMM @ICME18-19), tutorials (ACM-MM, CVPR, FG), PC (\eg CVPR, FG, MIRP, MMEDIA, AAAI, ICCV), reviewer (\eg TBioCAS, TIP, TPAMI), and President of IEEE@NEU and Rel. Officer of IEEE SAC R1 Region. Completed NSF REUs ('10 \& '11); co-op at Analogic Corp. ('12) BBN Tech. ('13); intern at MIT LL ('14), STR ('16-'17), Snap Inc. ('18), ISMConnect ('19). 

\end{IEEEbiography}
\vspace{-10mm}

\begin{IEEEbiography}[{\includegraphics[width=1in,height=1.2in,clip,keepaspectratio]{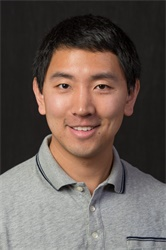}}]{Ming Shao}
Received the B.E. degree in computer science, the B.S. degree in applied mathematics, and the M.E. degree in computer science from Beihang University, Beijing, China, in 2006, 2007, and 2010, respectively. He received the Ph.D. degree in computer engineering from Northeastern University, Boston MA, 2016. He is a tenure-track Assistant Professor affiliated with College of Engineering at the University of Massachusetts Dartmouth since 2016 Fall. His current research interests include sparse modeling, low-rank matrix analysis, deep learning, and applied machine learning on social media analytics. He was the recipient of the Presidential Fellowship of State University of New York at Buffalo from 2010 to 2012, and the best paper award winner/candidate of IEEE ICDM 2011 Workshop on Large Scale Visual Analytics, and ICME 2014. He has served as the reviewers for many IEEE Transactions journals including TPAMI, TKDE, TNNLS, TIP, and TMM. He has also served on the program committee for the conferences including AAAI, IJCAI, and FG. He is the Associate Editor of SPIE Journal of Electronic Imaging, and IEEE Computational Intelligence Magazine. He is a member of IEEE.
\end{IEEEbiography}


\vspace{-10mm}
\begin{IEEEbiography}[{\includegraphics[width=1in,height=1.2in,clip,keepaspectratio]{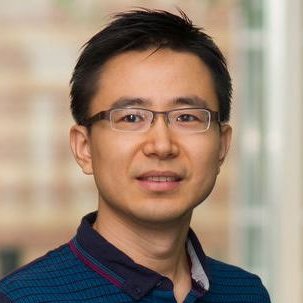}}]{Yun Fu}
(S’07-M’08-SM’11-F’19) received the B.Eng. degree in information engineering and the M.Eng. degree in pattern recognition and intelligence systems from Xi’an Jiaotong University, China, respectively, and the M.S. degree in statistics and the Ph.D. degree in electrical and computer engineering from the University of Illinois at Urbana-Champaign, respectively. He is an interdisciplinary faculty member affiliated with College of Engineering and the College of Computer and Information Science at Northeastern University since 2012. His research interests are Machine Learning, Computational Intelligence, Big Data Mining, Computer Vision, Pattern Recognition, and Cyber-Physical Systems. He has extensive publications in leading journals, books/book chapters and international conferences/workshops. He serves as associate editor, chairs, PC member and reviewer of many top journals and international conferences/workshops. He received seven Prestigious Young Investigator Awards from NAE, ONR, ARO, IEEE, INNS, UIUC, Grainger Foundation; eleven Best Paper Awards from IEEE, ACM, IAPR, SPIE, SIAM; many major Industrial Research Awards from Google, Samsung, Amazon, Konica Minolta, JP Morgan, Zebra, Adobe, and Mathworks, etc. He is currently an Associate Editor of the IEEE Transactions on Neural Networks and Leaning Systems. He is fellow of IEEE, IAPR, OSA and SPIE, a Lifetime Distinguished Member of ACM, Lifetime Member of AAAI, and Institute of Mathematical Statistics, member of Global Young Academy, INNS and Beckman Graduate Fellow during 2007-2008.
\end{IEEEbiography}




\end{document}

%% file: sections/1-introduction.tex
\IEEEPARstart{A}{bout} a decade ago, pioneers in visual kinship recognition research \color{review}published\color{black}~the seminar \color{review}work in detecting family relationships with face images~\cite{fang2010towards}. Let us now look back\color{black}~at this progress: reflect on the trends, successes, and failures of the past ten years. Furthermore, let us highlight the key challenges, practical use-cases, and promising future directions for research. By doing so, we provide a single resource to compare \ac{sota} methods on many tasks recorded and examined.

\subsection{So what? Who cares?}  
Kinship recognition has a multitude of practical and scholarly uses - relationships provide rich information in sociology, anthropology, and genetics;  privacy protections and concerns, along with potential use-cases that can be found in social media, personal discovery, entertainment, and more. Besides its entrepreneurial value, visual kinship recognition has significant non-commercial (or humane) value as well. For instance, in cases of missing children, reconnecting families split across refugee camps, border control and customs, criminal investigations, ancestral-based studies, and even genome-based research. Socially, family gives a sense of belonging (\ie membership, connection). Per Furstenberg,
\vspace{3mm}
\begin{quote}\small{
    ... important function of family systems receives far less attention in the literature than it merits: The family ... social arrangement responsible for giving its members a sense of identity and shared belonging ... not only those inside the natal family household but also among relations living elsewhere as well~\cite{furstenberg2020kinship}.}
\end{quote}

Hence, a recent surge in many seeking out their pedigree. With an abundance of visual data online, familial resources can benefit.

\begin{figure*}
\centering
    \includegraphics[width=\textwidth]{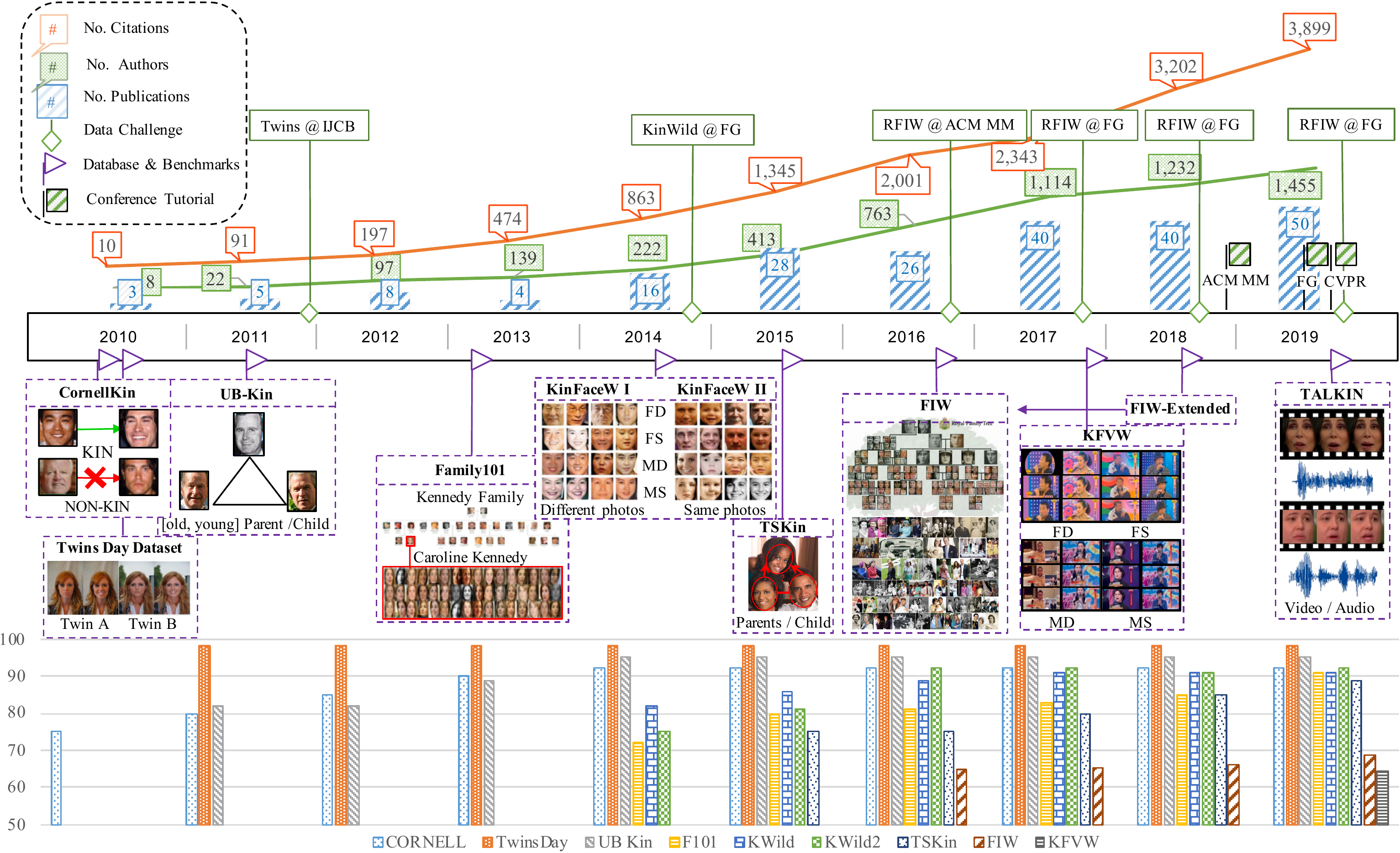}
    \caption{\small\textbf{The last decade of research in visual kinship recognition.} A timeline showing correlations between the data released (below timeline) and different citation metrics and events to indicate the impact throughout the research community (above timeline). To collect the statistics on citation-metrics, we built a pipeline to scrape the data needed from the web for the plots above: (1) \emph{Publish or Perish}~\cite{publishperish} was installed on a Mac Book Pro to scrape from various sources (\ie Google Scholar, Cross Ref, and Scorpus) and saved as CSV files; (2) CSV files were converted using Python; (3) the Mendeley application was then used to automatically merge duplicates, while keeping as much information as possible amongst the set; (4) queried Google Scholar for all \emph{Related Works} and \emph{Cited By} using PyPi's scholarly (\href{https://pypi.org/project/scholarly/}{https://pypi.org/project/scholarly/}). PyPi's scholarly both extended the paper-pile and increased the amount of metadata available (scholarly provides richer metadata per query, allowing us to fill in missing abstracts that are critical for the next step). Finally, we clustered the documents by abstract via \ac{tfidf}~\cite{ramos2003using}. The cluster made-up of a majority of papers on kinship recognition in multimedia: this reduced the burden of manual inspection of hundreds of thousands to thousands. It is important to note that only citation metrics were considered here, leaving out other factors of impact like the \emph{number of times tweeted}, \emph{Github stars}, and other indicators that may indicate research impact.}
    \label{fig:timeline}
\end{figure*}

\subsection{Purpose of this survey} 
We review the state of automatic kinship recognition after the first decade of research - with emphasis on the milestones that led us up to now (Fig.~\ref{fig:timeline}). Furthermore, we reflect on the problem statements of the different tasks to establish clear definitions and an understanding across the domain in a consistent manner for us to build on the upcoming decade. For this, we aim to use consistent terminology, assess the practical usefulness (or lack thereof), and highlight any outstanding challenges and obstacles that prevent the transition of visual kinship recognition technology from research to reality. With the problems made clear, and the practical significance properly measured, we compile a list of the \ac{sota} scores and methods of the main tasks with emphasis on the large-scale \ac{fiw}. All source code used to generate the scores is provided as supplemental, \ie scripts (run-specific), pipelines (task-specific), and tutorials (topic-specific).

In light of recent advances in the \ac{rfiw} challenge, and with the additional tasks added for the most recent 2020 edition, we review the details for the different paradigms of kin-based problems in the visual domain as formulated for the large-scale, multi-task \ac{fiw} database. We also look back at the existing datasets for visual kin-based problems motivated by different real-world scenarios (Table~\ref{tab:datasets}). In summary, our overarching goal is to set the stage for the next decade by clearly reporting experimental specifications, evaluation protocols, and the corresponding \ac{sota} methods supported with results, analysis, and scripts to reproduce the aforementioned.

There exists many unanswered questions that are potential future directions in machine vision research, like studies on familial and inheritance (\ie nature-based), and beyond. We take a glimpse at these promising next steps, while highlighting key challenges that we must overcome - both intrinsic to the image and inherent to the problem. We urge attempts in research to conduct cross-discipline studies as it is time to form such synergy.

One motivation of this survey is to establish cohesive views of the major milestones via protocols that are clearly defined, data splits that are ready for download, and trained models that make baselines reproducible.\footnote{\href{https://github.com/visionjo/pykinship}{https://github.com/visionjo/pykinship}} Hence, components to reproduce experiments that we report make up part of the supplemental material. We cover the edge cases that challenge \ac{sota}, including an examination of the different settings and training tricks that further our abilities in kin-based detection from faces.  

\begin{table*}[!t]
 \footnotesize\keepXColumns
 \renewcommand\theadfont{\bfseries\itshape}
 \centering
\scriptsize
    \caption{\textbf{Publicly available datasets for kinship recognition}. Each listed by the original name per reference. Kin-based image (or video) stats, which include the label types that support a specific metric to evaluate with and for which there exists a respective \ac{sota} to reference. URLs to the project page (\ie source for download) are included for each.  Abbreviations describing \emph{Stats} are for the number of families (\textbf{F}), face count (\textbf{f}), number of unique people (\textbf{P}), sample count (\textbf{S}), image count (\textbf{I})\color{review}, video count (\textbf{V}), and multimedia (\textbf{MM})\color{black}.}\label{tab:datasets}
    \begin{tabular}{r|>{\centering\arraybackslash}p{.3in}|>{\centering\arraybackslash}p{1.3in}|>{\centering\arraybackslash}p{1.5in}|>{\centering\arraybackslash}p{1.2in}|>{\noindent\justifying\arraybackslash}p{1.12in}}
        \textbf{DB}& \textbf{Ref(s)}& \textbf{Stats} & \textbf{Label types} & \textbf{Metric, performance, \ac{sota}} & \textbf{Web}\tabularnewline\midrule
        \multirow{2}{*}{CornellKin} 
        &\multirow{2}{*}{\cite{fang2010towards}}& \multirow{2}{*}{\textbullet~150~\textbf{F} \textbullet~300~\textbf{S} \textbullet~300~\textbf{f}} &\multirow{2}{*}{parent-child} &verification accuracy  94.4\%~\cite{kohli2018supervised} & \href{http://chenlab.ece.cornell.edu/projects/KinshipVerification/}{chenlab.ece.cornell.edu/ projects/KinshipVerification/}\tabularnewline \midrule
        
        \multirow{2}{*}{UB Face} &\multirow{2}{*}{\cite{shao2011genealogical}~\cite{Xia201144}}& \multirow{2}{*}{\textbullet~200~\textbf{F} \textbullet~250~\textbf{P} \textbullet~600~\textbf{f} \textbullet~400~\textbf{S}}& \multirow{2}{*}{([young, old] parent)-child} &\multirow{2}{*}{accuracy, 95.3\%~\cite{kohli2018supervised}} &  \href{http://www1.ece.neu.edu/~yunfu/research/Kinface/Kinface.htm}{www1.ece.neu.edu/yunfu/ research/Kinface/}\tabularnewline
        \midrule
        
        \multirow{2}{*}{Twins Day} &  \multirow{2}{*}{\cite{vijayan2011twins}}& \textbullet~1,736~(finger, 3D face, iris, DNA)~\textbf{S} \textbullet~197~\textbf{I} & \multirow{2}{*}{twin pairs}& \multirow{2}{*}{accuracy, 98.8\%~\cite{vijayan2011twins}}& \href{https://biic.wvu.edu/data-sets/twins-day-dataset-2010-1015}{/twins-day-dataset-2010-1015}\tabularnewline\midrule
    
        \multirow{2}{*}{SibFace} & \multirow{2}{*}{\cite{bottino2012new}} &\multirow{2}{*}{\textbullet~184~\textbf{S} \textbullet~78~\textbf{P} \textbullet~78~\textbf{F} \textbullet~184~\textbf{f}}& siblings (brothers, sisters, mixed)&\multirow{2}{*}{accuracy, 52.5\%~\cite{guo2014graph}}&
       \href{https://areeweb.polito.it/ricerca/cgvg/siblingsDB.html}{areeweb.polito.it/ricerca/ cgvg/siblingsDB.html}\tabularnewline\midrule

       \color{review}UvA-NEMO Smile\color{black}  &\cite{dibeklioglu2013like} &\textbullet~162~\textbf{P} \textbullet~515~\textbf{V} \textbullet~512~\textbf{S}& 7 relationships (core family)&  accuracy, 88.16\%~\cite{boutellaa2017kinship}& \href{https://www.uva-nemo.org/}{https://www.uva-nemo.org/}\tabularnewline\midrule
        \multirow{2}{*}{Family101} &\multirow{2}{*}{\cite{fang2013kinship}} &\multirow{2}{*}{\textbullet~101~\textbf{F} \textbullet~607~\textbf{S}}& \multirow{2}{*}{family-tree structure}&  \multirow{2}{*}{rank@10, 70.1\%~\cite{wu2018kinship}}& \href{http://chenlab.ece.cornell.edu/projects/KinshipClassification/index.html}{chenlab.ece.cornell.edu/ projects/KinClassification/}\tabularnewline\midrule

        \multirow{1}{*}{KFW I + II} &\multirow{1}{*}{\cite{lu2015fg}}&\textbullet~533 + 1,000~\textbf{P} \textbullet~1,066+2,000~\textbf{f} & parent-child; same + different photo& \multirow{1}{*}{accuracy, 96.9\% + 97.1\%~\cite{kohli2018supervised}} &\multirow{1}{*}{\href{http://www.kinfacew.com/}{http://www.kinfacew.com/}}\tabularnewline\midrule

        \multirow{1}{*}{TSKin} &  \multirow{1}{*}{\cite{qin2015tri}}&\multirow{1}{*}{\textbullet~787~\textbf{F} \textbullet~2,589~\textbf{S}} &\multirow{1}{*}{(father \& mother)-child}& \multirow{1}{*}{accuracy, 91.4\%~\cite{liang2018weighted}}&
        \href{http://parnec.nuaa.edu.cn/xtan/data/TSKinFace.html}{parnec.nuaa.edu.cn/TSKinFace} \tabularnewline\midrule

        \multirow{2}{*}{FIW}& \cite{ robinson2018visual, robinson2020recognizing} &\textbullet~1,000~\textbf{F} \textbullet~33,000~\textbf{f} \textbullet~1-M~\textbf{P} \textbullet12,000~\textbf{S} \textbullet~13,000~\textbf{I} &large-scale; person-, family-, and image-level& accuracy, 78\%; tri-subject accuracy, 79\%; mAP 18\% \& rank@5 60\%~\cite{robinson2020recognizing} &\href{https://web.northeastern.edu/smilelab/fiw/}{https://web.northeastern.edu/
        smilelab/fiw/}\tabularnewline\midrule
        
        
        \multirow{2}{*}{KFVW} &\multirow{2}{*}{\cite{yan2018video}} & \textbullet~418 (video) \textbf{P} (100-500 \textbf{f} per video);&  \multirow{2}{*}{parent-child} &
         \multirow{2}{*}{accuracy, 61.8\%~\cite{yan2018video}}& \multirow{2}{*}{\href{https://www.kinfacew.com/index.html}{https://www.kinfacew.com}}\tabularnewline\midrule

        \multirow{2}{*}{KIVI} &    \multirow{2}{*}{\cite{kohli2018supervised}} & \multirow{2}{*}{\textbullet~503 (video) \textbf{S} \textbullet~503 \textbf{I}} & \multirow{2}{*}{7 relationships (core family)} &   \multirow{2}{*}{accuracy, 83.2\%~\cite{kohli2018supervised}} & \href{http://iab-rubric.org/resources/KIVI.html}{http://iab-rubric.org/resources/KIVI.html} \tabularnewline\midrule

        \color{review} \multirow{2}{*}{FIW-MM} &    \multirow{2}{*}{\cite{robinson2020familiesinMM}} & \multirow{2}{*}{FIW + 937 \textbf{MM}} & {FIW + multimedia for $\approx$~200~\textbf{F} (\ie video, audio, and contextual data)} &   \multirow{2}{*}{EER,  89.8\%; mAP, 0.24~\cite{robinson2020familiesinMM}} & \href{https://web.northeastern.edu/smilelab/fiw/}{https://web.northeastern.edu/
        smilelab/fiw/}\color{black} \tabularnewline
        
    \end{tabular}
\end{table*}

\subsection{Related surveys}
The first survey on visual kinship recognition gave an overview of the \ac{sota} methods and data resources of the time~\cite{wu2016kinship}. The authors proposed future directions with great emphasis on the lack of labeled data both in sample counts and relationship label types. Hence, Wu~\etal claim that a \ac{cnn} was inferior to the traditional metric-based proved to hold true~\cite{wu2016kinship}. About the same time came the release of the large-scale \ac{fiw} dataset to support contemporary data-driven solutions~\cite{robinson2016families}. Georgopoulos~\etal then surveyed kinship and age in \ac{fr}~\cite{georgopoulos2018modeling}. Although a comprehensive piece, kin-based problems ought to be surveyed independently. Nonetheless, prior knowledge of one could benefit the other; knowledge of various soft biometrics tends to complement and are beneficial (\eg gender and emotion). On the one hand, a survey on age or kinship should mention the other; however, the directed graphs and concepts of inheritance make kin-based studies worthy of surveying as an independent topic. \color{review}Finally, looking at the problem from the view of understanding age, we make a similar claim-- knowledge of kinship could most certainly help an age estimate. For instance, we have photos of a father at one or more known ages, while we are tasked to predict the age of the son. The knowledge available in the set of faces of known age is available as a prior and be modeled accordingly. Point is, we do not mean that the different attribute-based face understanding tasks ought to be treated as independent. Note, we do claim a study on the modeling and analysis of kin-based media should be surveyed alone. Still, as we cover later, age, gender, and variations of in other attributes do provide additional challenges in kin-based tasks.\color{black}

Most recently, Qin~\etal surveyed kinship recognition methods as being founded on \emph{a measure of kinship traits} or \emph{statistical learning}~\cite{qin2019literature}. Furthermore, the groups were characterized for being \emph{low}\ or \emph{mid-level features}, \emph{metric learning}, or \emph{transfer learning}. The authors reported scores for several kin-based datasets. Additionally, \emph{human} performance compared to machines was included. As part of their work, the authors proposed a standardized vision system based on four-steps to provide a generic, modular solution. To complement this, we define the problems consistently for the many kin-based tasks, and with details on the \ac{sota} for each.


\subsection{Organization} We first look at the major-milestones for the first decade of research in visual kinship recognition. \color{review}For this, we review the problem as it evolved over time, along with the public data supporting the progress, and with data statistics and web links of the source. \color{review2}Furthermore, we look at kinship recognition research that compares humans to machines, showing resemblance is detectable via the human eye (Section~\ref{sec:background}). \color{black}Next, we introduce kin-based \color{review}tasks by discussing the different problem statements (Section~\ref{sec:data:benchmarks:resources}). 
\color{review2}Following this, we discuss experimental details for each of the tasks-- summarize the protocols of the laboratory-based evaluations, including the data splits, metrics, and baseline results for each (Section~\ref{sec:sota:visualkin}).  Then, we cover methodologies, both traditional and deep learning based for both discriminative and generative (Section~\ref{sec:methods}).
\color{review}Then, we discuss technical challenges preventing this technology from working reliably in real-world applications. Specifically, we cover the current limitations of \ac{sota}-- raise the discussion on a more broad perspective of the impact from kin-based technologies (\ie in our everyday lives and as our capabilities are enhanced. This is supported by a rigorous analysis on the edge cases and commonalities of falsely predicted samples. We highlight challenges posed by nature and the environment, and then shine a light on the inherent difficulties of obtaining sufficient data for kin-based problem (Section~\ref{sec:challenges}). \color{black}This leads to the applications that line up with specific task-evaluations, both existing (\ie practically existing) and high in potential (\ie hypothetically possible). Emphasis is especially placed on the more robust models-- typically, assuming we can improve the performance of the current \ac{sota} (Section~\ref{sec:applications}). We close on the broader impacts, followed by our forecast for both short and long-term directions and problems for which automatic kinship understanding may take. Finally, we conclude (Section~\ref{sec:discussion}).

%% file: sections/2-background.tex
\section{Background Information}\label{sec:background}
The story of visual kinship recognition research can be told through the data. Therefore, we speak of the progress through the first decade from the perspective of the resources available (Fig.~\ref{fig:timeline}), and it is shown that interest has been contingent on the amount and quality of labeled data. We end by discussing the data challenges, workshops, and tutorials used to motivate researchers.

\subsection{The evolution of the problem}
An increasing number of researchers have focused their attention on the problem of learning families in photos since the seminal paper was published in 2010~\cite{fang2010towards}. \color{review}The research progress had the past decade coincided with the supporting labeled data released in part to it. Following Fig.~\ref{fig:timeline}, we will next look back at the problem.

\color{review}A trend observed in the progress in visual kinship recognition over the past decade is its correlation with the respective data resources released for public use. Critical points in the research stemmed from the respective problem statements supported by data labeled for the task. Hence, to review the problem statements and protocols as the problem evolved over time.\color{black}

\color{review}
Fang \etal proposed training machinery to visually discriminate between \emph{KIN} and \emph{NON-KIN} using various facial cues~\cite{fang2010towards}. Specifically, the authors demonstrated an ability to verify kinship given a face pair. To support this, they built and released the first facial image-based kinship database called Cornell Kin. Cornell Kin consisted of 150 face pairs of type \emph{parent}-\emph{child} from the web (\ie public figures, politicians, and other famous persons). Next came biometric data of twins collected at an annual event called~\emph{Twins' Day}~\cite{vijayan2011twins}. This effort yielded in a collection of 197 individuals of multiple modalities (\ie finger prints, 3D face scans, images of irises, and DNA samples) spanning multiple years (\ie from 2011 onward, each year new samples for subjects were added). Shao~\etal then proposed UB Face made-up of 250 parent-child, each supported by three samples (\ie child, parent at a younger age, and parent at an older)~\cite{shao2011genealogical}. The motivation for the pairs having a sample of each parent at a young and old age was directly spawned up from consideration for the difficulty imposed by large age gaps~\cite{xia2012understanding}. Soon thereafter came Family101~\cite{fang2013kinship}-- the first image collection with knowledge of family tree information, with 101 trees and multiple samples per subject. In 2014, Kin-wild I \& II then provided a rich collection of 2,000 parent-child pairs~\cite{lu2014kinship}-- will be discussed in the following section, along with Section~\ref{subsec:verification}, describing this database had a significant impact for many expert researchers who proposed clever metric learning methods. Following this came the \ac{tsk} dataset, which structured the problem differently: given a parents-child pair (\ie both parents and a child), determine \emph{KIN} or \emph{NON-KIN}. Then, came \ac{fiw}, which remains the largest kin-based image collection up to today. \ac{fiw} is the main data used for experiments in this survey-- more information provided in detail in the sections to come. Finally, and most recently, was the release of multimedia collections in support of kin-based tasks. First of these was released in 2017 - \ac{kfw} released video data for parent-child pairs, which then allowed for richer, dynamic models to be trained across video frames. Last year, in 2019, came the release of kin-based data that also leveraged audio media, \ie \ac{talkin}. Lastly, \ac{fiw} was extended with multimedia data added to over 200 of its 1,000 families~\cite{robinson2020familiesinMM}. Specs of the aforementioned data (\eg label types, \ac{sota}, reference links) are in Table~\ref{tab:datasets}. Furthermore, the advancements in methodologies are later covered in detail.

\color{black}


\begin{figure}[!t]
    \centering
    \includegraphics[width=\linewidth]{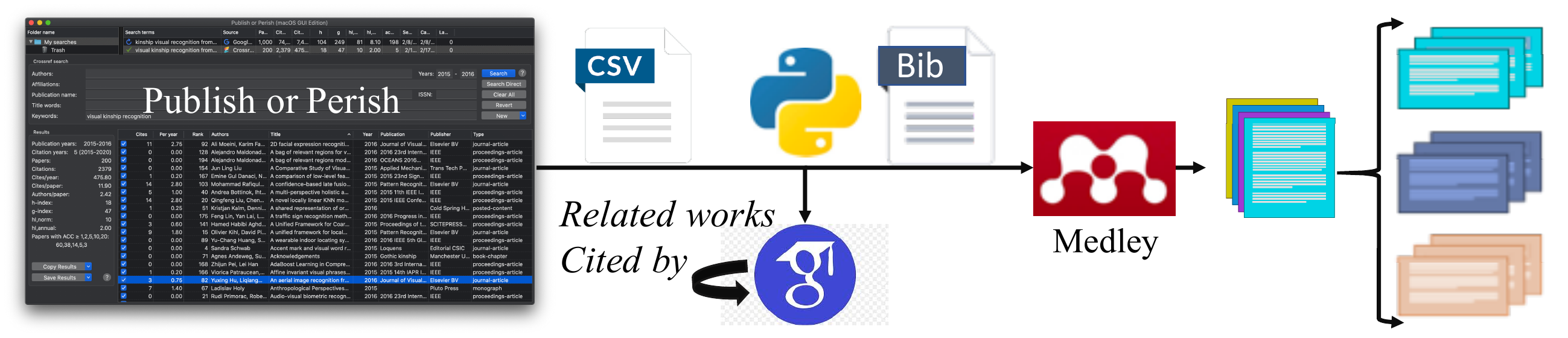}
    \caption{\textbf{Workflow to scrape publication metadata for Fig.~\ref{fig:timeline}.} From \emph{Publish or Perish}~\cite{publishperish}, we queried Scholar for \emph{Related works} and \emph{Cited by}, increasing the size of our list nearly 20-fold. Mendleley merged duplicates, while keeping as much information as possible. Applied \ac{nlp} to cluster relevant documents.}
    \label{fig:citations}
\end{figure}

\color{review}To build Fig.~\ref{fig:timeline},\color{black}~\emph{Publish and Perish} was used to acquire the paper-pile for the analysis (Fig~\ref{fig:citations}). For this, a series of queries was executed, each using \emph{visual kinship recognition} as the keywords: Google Scholar, limited to 1,000 search results per query, was run two times (\ie 2010-2020 and 2015-2020); Scorpus was queried from 2010-2020, as only 48 items were found; CrossRef, with a limit of 200 items per search, was queried by year of publication (\ie 2010-11, 2011-12, \dots, 2015-16, 2016, 2017, \dots, 2020). Notice that the years were set such that fewest were expected the first year, more in the first half of the decade, and the most in the latter half. Many papers returned were not on automatic kinship recognition in visual media. However, using the \ac{tfidf} representation, we were able to quickly filter out irrelevant papers by semi-supervised clustering (\ie side-information-based) cosine-similarity k-means~\cite{robinson2018visual} with labels assumed positive for the papers with keywords or titles that contain \emph{visual kinship recognition}.~\color{review}A rise in the number of annual publications indicates an increase in interest of researchers; the impact on the research community, as a whole, nearly grows exponentially (\ie citation count). The incentive provided by data challenges, along with the increase in labeled data, influence the attention given to kin-based problems in multimedia (MM).\color{black}

\color{review2}
\subsection{Humans recognizing kinship in photos}
\color{review}
Several works in vision research evaluate the ability of humans to detect kinship given a face pair. In the seminar work, Fang~\etal evaluated humans using a subset of their Cornell Kin dataset and found that, on average, humans were about 4\% worse than the machinery: average human performance  was 67.19\%, with the top performance reaching 90\% and the worst at just 50\%~\cite{fang2010towards}. The authors also found supporting results of a previous cognitive study on the perceivable similarity of offspring of different genders (\ie sons tend to be more recognizable as kin than daughters). Provided the time of this work- a time when it was still unclear whether \emph{KIN} from \emph{NON-KIN} signals are detectable via facial cues- this contribution was not only to compare humans and machines but was to get a sense if even possible for humans.\color{black}

\begin{figure*}[t!] 
	\centering
	\includegraphics[width=\textwidth]{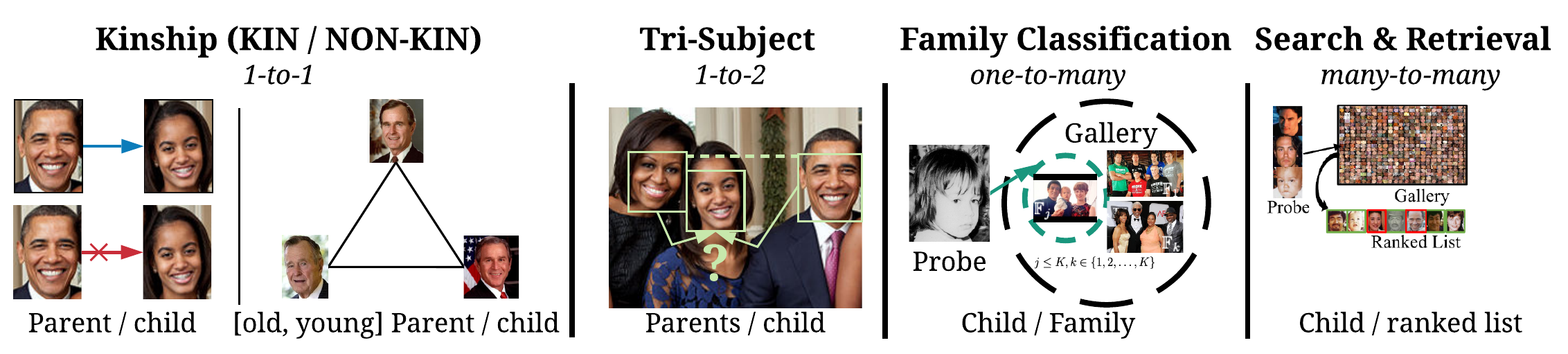}
	\caption{\textbf{Visual kin-based discriminate tasks.} Depicted are problems of verification (\ie one-to-one) and family classification (\ie one-to-many), along with the more recently supported tri-subject verification  (\ie one-to-two) and search \& retrieval for ``missing'' children (\ie many-to-many). The \ac{fiw} database is one of the major benchmarks that support each of these tasks as part of the annual data challenge \ac{rfiw}-- the most recent challenge supported three of these tasks, as family classification was found to carry less potential for practical use-cases. The data splits (\ie train, val, and test with no family overlap), protocols, and benchmarks are described in~\cite{robinson2020recognizing}. Also, other data resources in Fig.~\ref{fig:timeline} and Table~\ref{tab:datasets} also support one or more of these tasks. Best viewed electronically.}
	\label{fig:tasks}
	\vspace{-3mm}
\end{figure*}

\color{review}The architects of \ac{fiw} re-evaluated humans at a greater scale, and with their more diverse data resource~\cite{robinson2018visual}. \color{black}Specifically, eleven relationship types, opposed to the four of Cornell Kin, were involved, and at pair counts ten-times that organized in 2010. \color{review}{The human evaluation found the following: not only were human's machines outperforming humans on the unconstrained data, but the authors showed no difference in performance with knowledge of the pairwise type previously given.}

More recently, a study focused on comparing human performance verifying kinship on a grander scale. Furthermore, the study presented extended coverage in the topic of human performance from the view of psychology. That is the work done by Lopez~\etal~\cite{lopez2018kinship}. Their evaluation displayed a face pair and prompted over 300 individuals connected through crowdsourcing services to answer: \emph{Are these two people related (\ie part of the same family)}, with possible responses set as \emph{Yes} or \emph{No}. Face pairs were made up of possible and negatives from both Kin-Wild~\cite{lu2015fg} (\ie image set) and Uva Nemo Smile~\cite{dibeklioglu2013like} (\ie video set)~\cite{dibekliouglu2012you}. The machine again outperformed the human. Taking it a step further, Hettiachchi~\etal analyzed the effects of gender and race, in both the human and data, finding that both genders are similar in ability to recognize kin, while both tend to perform best on same gender pairs (\eg brother-brother, mother-daughter, \etc)~\cite{hettiachchi2020augmenting}. Furthermore, the authors validated previous findings in own-race bias for humans recognizing kinship.
\color{black}




\subsection{Data challenges and incentives}
Challenges date back to 2011, where multi-modal data for twins was collected annually and in a highly controlled setting (\ie \emph{Twins Day}~\cite{vijayan2011twins}). Also, starting in 2014 were data challenges on unconstrained face data~\cite{lu2014kinship}. Then, Lu~\etal attracted many with a \ac{fg} challenge with \ac{kfw}~\cite{lu2015fg}. Robinson~\etal expanded the data challenges as part of a 2017 ACM MM Workshop using the first large-scale visual kinship recognition dataset~\cite{robinson2017recognizing}, which was followed by three consecutive \ac{fg} challenges - an annual effort that still occurs nowadays~\cite{robinson2020recognizing} (\ie 2018-2020\color{review}, with 2020 still accessible via Codalab\footnote{\href{https://competitions.codalab.org/competitions/21843}{https://competitions.codalab.org/competitions/21843}}\color{black}). Besides, over five hundred teams partook in \ac{rfiw} on Kaggle.\footnote{\href{https://www.kaggle.com/c/Recognizing-Faces-in-the-Wild}{https://www.kaggle.com/c/Recognizing-Faces-in-the-Wild}} Recently, there have been several tutorials at top-tier conferences (\ie ACM MM18~\cite{robinson2018recognize}, CVPR 2019\footnote{\href{https://web.northeastern.edu/smilelab/fiw/cvpr19_tutorial/}{https://web.northeastern.edu/smilelab/fiw/cvpr19\_tutorial/}}, and \ac{fg} 2019\footnote{\href{http://fg2019.org/participate/workshops-and-tutorials/visual-recognition-of-families-in-the-wild/}{http://fg2019.org/visual-recognition-of-families-in-the-wild}}). \color{review2}The human evaluations were done using volunteers in a non-competitive forum.\color{black}


%% file: sections/experimental.tex
\begin{table*}[t!]
    \centering
    \caption{\textbf{T1 Counts.} Number of unique pairs (\textbf{P}), families (\textbf{F}), and face samples (\textbf{S}), with an increase in counts and types since~\cite{robinson2017recognizing}.}
    \begin{tabular}{p{.1in}m{.1in}|m{.29in}m{.29in}m{.29in}|m{.29in}m{.29in}m{.29in}m{.29in}|m{.29in}m{.29in}m{.29in}m{.30in}|m{.29in}}
        & & \textbf{BB}& \textbf{SS}& \textbf{SIBS}& \textbf{FD} & \textbf{FS} & \textbf{MD} & \textbf{MS} & \textbf{GFGD} & \textbf{GFGS} & \textbf{GMGD} & \textbf{GMGS}& \textbf{Total}\\\hline
     \parbox[t]{2mm}{\multirow{3}{*}{\rotatebox[origin=c]{90}{\emph{Train}}}}&\textbf{P} & 991  & 1,029 &1,588 & 712 & 721& 736& 716 & 136 & 124 & 116 & 114 &6,983\\
    \multirow{3}{*}{} &\textbf{F}  &303 & 304 & 286 & 401 & 404 & 399 & 402 & 81 & 73&71 & 66 &2790\\
    \multirow{3}{*}{} &\textbf{S} &39,608& 27,844 & 35,337& 30,746  &46,583 & 29,778&  46,969& 2,003 &  2,097  &1,741 & 1,834  &264,540\\\hline
    
    \parbox[t]{2mm}{\multirow{3}{*}{\rotatebox[origin=c]{90}{\emph{Val}}}} &\textbf{P}  & 433 & 433 & 206& 220 & 261 & 200 & 234 & 53 & 48 & 56 & 42 & 2,186 \\
    
    \multirow{3}{*}{} &\textbf{F}  &74  & 57& 90 & 134& 135& 124& 130& 32& 29& 36&27 &868\\
    \multirow{3}{*}{} &\textbf{S}  & 8,340 & 5,982 & 21,204& 7,575 &9,399&8,441 &7,587 & 762 &879 & 714 & 701 & 71,584\\\hline

    \parbox[t]{2mm}{\multirow{3}{*}{\rotatebox[origin=c]{90}{\emph{Test}}}} &\textbf{P}  &  469& 469 & 217 & 202& 257 & 230 & 237 & 40 & 31 & 36 & 33&2,221 \\
    \multirow{3}{*}{} &\textbf{F}  & 149  & 150  & 89 & 126 & 133 & 136 & 132 & 22 & 21 & 20 & 22 & 1,190\\
    \multirow{3}{*}{} &\textbf{S}  & 3,459 &2,956 &967 &3,019&3,273&3,184& 2,660 &121&96&71&84&39,743\\
    
    \end{tabular}\label{tbl:track1:counts} 
\end{table*}

\color{review}
\section{Experimental}\label{sec:sota:visualkin}

The organization of this section is as follows. First, we examine studies involving a human's ability to recognize kinship as imagery. Thus, deeming the soft-attribute of kinship as being detectable by the eye. Next, we review kin-based task protocols - each complete with a problem statement, data splits, metrics, and baseline solutions. We then highlight commonalities in problem formulation and proposed solutions for the various tasks. Following this, we describe traditional and deep solutions. We then put this in perspective with the \ac{rfiw} data challenge series - four editions (\ie 2017~\cite{robinson2017recognizing}-2020~\cite{robinson2020recognizing} and Kaggle Competition\footnote{\href{https://www.kaggle.com/c/recognizing-faces-in-the-wild}{https://www.kaggle.com/c/recognizing-faces-in-the-wild}} held just prior to the 2020 \ac{rfiw}). Finally, we discuss recent attempts to predict the appearance of family members' faces.

\color{black}

\subsection{Task Evaluations, Protocols, Benchmarks}\label{sec:task:eval}
The different kin-based tasks are separately described next. Specifically, the problem statements and motivations, the data splits and protocols, and, finally, the baseline experiments. We first cover the common settings prior to covering settings unique to each of the tasks that follow in separate subsections.

\color{review}
\subsubsection{Common settings}\color{black}
The task-specific experiments were done using the \ac{fiw} dataset, for \ac{fiw} provides the most comprehensive set of face pairs for kin-based face recognition to date. Specifically, \ac{fiw} contains the data required to support modern-day data-driven deep models~\cite{duan2017advnet, li2017kinnet, wang2017kinship}. \color{review}With over 12,000 family photos for 1,000 disjoint family trees the data contains various counts for faces, samples, members, and relationships per family (Fig~\ref{fig:gronk:montage}). Hence, the faces of the image collection were cropped and organized per family-- family members have at least one face sample, though most have many. The \ac{fiw} dataset consists of three sets with no overlap in family or identity.-- \emph{train}, \emph{val}, and \emph{test}. Specifically, 60\% of the families were assigned to the \emph{train} set, 20\% to \emph{val}, and the remaining 20\% to \emph{test}. The test set remains ``blind'', as scoring is handled automatically for submissions to the respective codalab competitions. Also, the splits are consistent across different tasks.\color{black}


\begin{table*}[t!]
\centering
\caption{\color{review}\textbf{KinWild benchmarks.} Results for KinWild I and II.}
\label{tab:kinwild}
\centering
    \begin{tabular}{r|ccccc|ccccc}
    
     &\multicolumn{5}{c}{\textit{KinWild I}}& \multicolumn{5}{c}{\textit{KinWild II}} \\
     & \textbf{FD}& \textbf{FS}  & \textbf{MD} & \textbf{MS}   & \textbf{Avg.}& \textbf{FD}& \textbf{FS}  & \textbf{MD} & \textbf{MS}   & \textbf{Avg.} \\\midrule
     
 MNRML~\cite{lu2014neighborhood}   & 72.5&66.5& 66.2 & 72.0   & 69.9  & 76.9 & 74.3 & 77.4 & 77.0 & 76.4 \\
   PDFL~\cite{yan2014prototype} & 73.5& 67.5 &66.1 &73.1&70.0& 77.3& 74.7& 77.8 &78.0 & 77.0 \\
   DMML~\cite{yan2014discriminative} & 74.5 & 69.5& 69.5& 75.5&72.3& 78.5& 76.5& 78.5& 79.5 & 78.3\\
   multiviewSSL~\cite{ZHOU2016136} & 82.8 & 75.4 & 72.6 & 81.3 &78.0 & 81.8 & 74.0 & 75.3 & 72.5&75.9 \\
   SSML~\cite{fang2016sparse}& 81.7 &75.3 &71.4 &77.9&79.6& 82.4& 78.6& 79.8& 77.9 & 79.7\\
   SPML-P~\cite{8019375}&75.4& 84.3& 81.1 &72.4&78.3& 82.4& 77.6& 76.6& 76.2&78.2\\
   ELM~\cite{wuX2018kinship} & 70.0& 64.2& 73.0& 77.2&71.1& 78.6 &73.6 &81.0 &79.6 &78.2\\
   KVRL-fcDBN~\cite{kohli2016hierarchical} & \textbf{96.3}&\textbf{98.1} & \textbf{98.4}& \textbf{90.5} & \textbf{96.1} & \textbf{94.0} & \textbf{96.0} & \textbf{96.8} & \textbf{97.2} & \textbf{96.2}\\
   MvGMML~\cite{hu2019multi}&69.3& 73.1& 69.4 &72.8 &71.1& 70.4 &73.4 &65.8& 69.2 &69.7\\
   DDMML~\cite{lu2017discriminative}&79.1&86.4&87.0&81.4 & 83.5 & 83.8 & 87.4 & 83.0 & 83.2 & 84.3\\
   KML~\cite{zhou2019learning}&-&-&-&-&82.8 &- &- &- &-& 85.7\\
   MSIDA+WCCN~\cite{laiadi2020multiview} & 86.0 & 85.93& 90.1 &88.6 &87.7 &89.4 &82.8& 87.8 &88.0 &87.0\\
    \end{tabular}
\end{table*}

All faces from each set were encoded using ArcFace \ac{cnn}~\cite{wang2018additive}  (\ie 512 D), and with all pre-processing and training steps as described in the original work.\footnote{\href{https://github.com/ZhaoJ9014/face.evoLVe.PyTorch}{https://github.com/ZhaoJ9014/face.evoLVe.PyTorch}} Also common throughout was the metric to determine closeness: cosine similarity~\cite{nguyen2010cosine} measured the similarity between a pair of facial features $p_1$ and $p_2$ via
$$
\text{CS}(\pmb p_1, \pmb p_2) = \frac {\pmb p_1 \cdot \pmb p_2}{||\pmb p_1|| \cdot ||\pmb p_2||}.
$$


Scores were then either transformed to decisions upon being compared to threshold $\gamma$ (\ie $\text{CS}(p_1, p_2) > \gamma$ translates to \emph{KIN}, else, \emph{NON-KIN}) or were sorted by score (\ie ranked list). 



\subsubsection{Kinship verification \color{review}(T1)\color{black}}\label{sec:kinver}

The task of kinship verification is to determine whether or not a face pair are blood relatives. This classical Boolean problem has two possible outcomes, \emph{KIN} or \emph{NON-KIN}. Hence, the goal is to learning a \textit{one-to-one} mapping that transforms a pair of faces to one of two possible outcomes. Furthermore, the labels can be further split by the type of kin relation hypothesized for the given pair of faces-- convention measures accuracy per type and then a weighted average~\cite{robinson2018recognize}.

Before \ac{fiw}, vision researchers typically considered parent-child pairs, \ie \ac{fd}, \ac{fs}, \ac{md}, \ac{ms}. Also, some, though less, attention was given to sibling pairs, \ie \ac{ss}, \ac{bb}, and \ac{sibs}. Earlier research findings in psychology and computer vision found that each relationship type tends to share specific familial features~\cite{shao2011genealogical}. From this, most often a model per relationship type is trained, meaning the assumption that the type is known prior is made. Thus, the scope of types targeted is the scope that predictive power is assumed: additional kinship types would advance our understanding and capabilities of automatic kinship recognition. With \ac{fiw}, the number of facial pairs accessible for kinship verification has increased with the addition of grandparent-grandchildren types, \ie \ac{gfgd}, \ac{gfgs}, \ac{gmgd}, \ac{gmgs}.

\vspace{1mm}\noindent\textbf{Data splits.}
\color{review}
The two datasets used, \ac{kfw} and \ac{fiw}, follow the same settings. Both provide a list of pairs labeled as \emph{KIN} and \emph{NON-KIN}. The differences are in the number of pair types (\ie the addition of \emph{grandparent}-\emph{grandchild}) and the number of instances per data split. Data specifications are in Table~\ref{tab:datasets}.

\ac{kfw} provides two sets (\ie \ac{kfw} I \& II) and the four parent-child pair types. \ac{fiw} spans eleven different relationship types - the types used in 2020 \ac{rfiw} (Table~\ref{tbl:track1:counts}). The {\emph test} set is made up of an equal number of positive and negative pairs and with no family (and, hence, no identity) overlap between sets. 
\color{black}

\vspace{1mm}\noindent\textbf{Settings and metrics.}\label{subsec:track1:settings}

\begin{figure}[t!]
    \centering
    \includegraphics[width =1\linewidth]{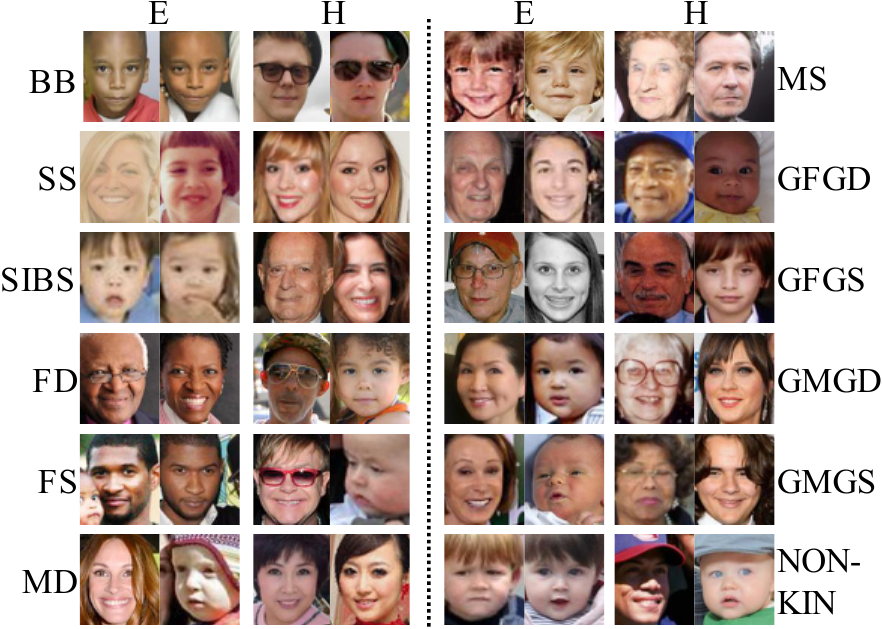}
    \caption{\textbf{T1 Sample pairs~\cite{robinson2020recognizing}.} Sample pairs with similarity scores near the threshold (\ie hard (H) samples), along with highly confident predictions (\ie easy (E) samples) in verification task.}
    \label{fig:track1:samples}
    \vspace{1mm}
\end{figure}

Verification accuracy is used to evaluate. Specifically,

$$
\text{Accuracy}_j = \frac{\text{\# correct predictions for j-th type}}{\text{Total \# of pairs for j-th type}},
$$
where \color{review}$j\in\{$4 relationship types and $\O$ for \ac{kfw} and 11 relationship types and $\O$ for \ac{fiw}$\}$ (listed in Fig.~\ref{fig:track1:samples}). \color{black}Then, the overall accuracy is calculated as a weighted sum (\ie weight by the pair count to determine the average accuracy).

\vspace{1mm}\noindent\textbf{Baseline and results.}
For all pair-types, the threshold was set to the value that maximizes the accuracy on the \color{review}\emph{val} set. The results on \ac{kfw} I and II are show in Table~\ref{tab:kinwild}. The results for \ac{fiw} are in Table~\ref{tab:benchmark:track1}, with \color{black}sample pairs that either 100\% or 20\% of all teams got correct are shown in Fig.~\ref{fig:track1:samples:submitted}.

\subsubsection{Tri-Subject verification \color{review}(T2)\color{black}}\label{sec:trisubject}
Tri-Subject Verification (T2) focuses on a different view of kinship verification-- the goal is to decide if a child is related to a pair of parents. This is believed to be a more realistic assumption, as knowledge of one parent typically means that the other parent can be inferred (\eg via marriage or birth records).


\vspace{1mm}\noindent\textbf{Data splits.} Following \cite{qin2015tri}, we create positive triplets by matching each father-mother pair with their true child. Then, we hold the parents constant and shuffle the children to generate the list of negative triplets. Counts per triplet type are listed in Table~\ref{tbl:track2:counts}.


Note that the number of potential negative samples is much greater than that of the positives. To avoid problems of having severely over-represented identities and families due to an abundance of face samples, the list of positive samples were provided weights that inversely represented the count per individual and family. This allowed for the positives to include an exhaustive list of all triplets, while smoothing out the negative sample distribution to be fair across family and subject. 
Specifically, the number of samples for any triplet $(F, M, C)$ was set to 5 (\ie better balance in positive set per sample), while the use of any $(F, M)$ parent-pair was set to 15 (\ie better balance of negative per sample), and then finally limiting the total number of triplets per family to 30 (\ie better balance of family in total). Again, the \emph{test} set was made to have an equal number of positive and negative. 
%

\vspace{1mm}\noindent\textbf{Settings and metrics.} Convention for face verification was again followed (\ie as described in for task 1 in Section~\ref{subsec:track1:settings}). Similarly, the verification accuracy was used to determine the performance for a set of results, which is first calculated per triplet-pair type (\ie FMD and FMS) and the weighted sum (\ie average accuracy) was the final score used to place in the leader-board.

\begin{figure}[t!]
    \centering
    \includegraphics[width =.9\linewidth]{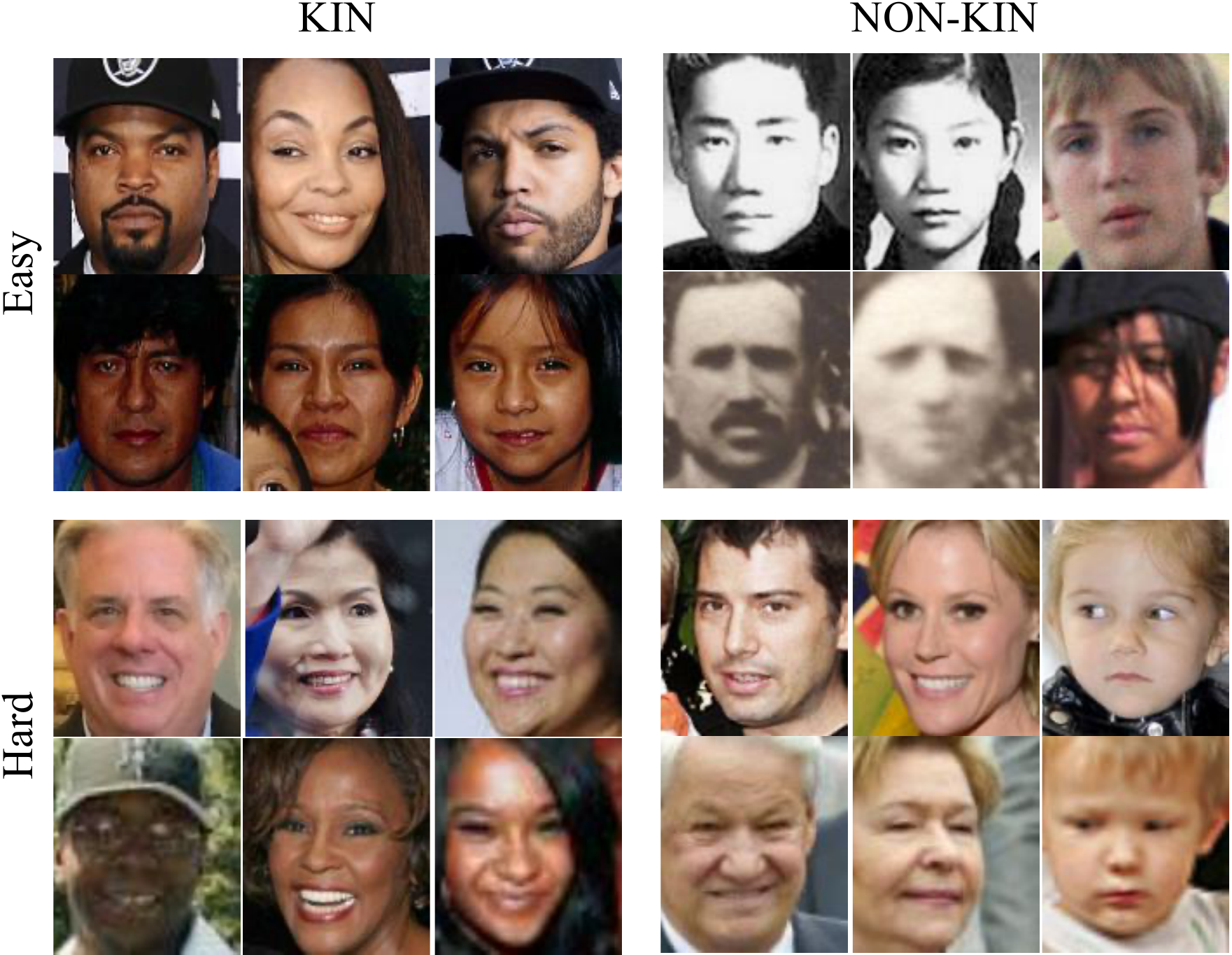}
    \caption{\textbf{Triplets with extreme scores (\ie true and false)~\cite{robinson2020recognizing}.} Each show FMS (top rows) and FMD (bottom) for tri-subject (T2).}
    \label{fig:track2:montage}
    \vspace{-1mm}
\end{figure}

\vspace{1mm}\noindent\textbf{Baseline and results.}
Baseline results are shown in Table~\ref{tab:benchmark:track2}, with samples of easier and more challenging samples for both \emph{KIN} and \emph{NON-KIN} triplets in Fig.~\ref{fig:track2:montage} and \ref{fig:track2:samples:submitted}. A score was assigned to the $ith$ triplet $(F_i, M_i, C_i)$ in the validation and \emph{test} sets using
$$\text{Score}_{i} =  avg(\cos{(F_i, C_i)}, \cos{(M_i, C_i)}),$$
where $F_i$, $M_i$ and $C_i$ are the face encodings of the father, mother, and child, respectively. 
Like done for verification, scores pass through a threshold $\gamma$ to transform the score to a label (\ie predict \emph{KIN} if the score was above the threshold; else, \emph{NON-KIN}). 
Like in T1, $\gamma$ was determined experimentally on \emph{val} and used for \emph{test}.

\subsubsection{Search and retrieval \color{review}(T3)\color{black}}\label{sec:search}
Search \& retrieval (T3) is a newer task posed as a \textit{many-to-many} paradigm, with all subjects represented by one-to-many samples. For this, template-based evaluations are imitated: subject to search against given as the \emph{probe}, with the \emph{gallery} of subjects with family labels as the ground-truth. Thus, the goal of the task is to rank all subjects in the \emph{gallery} such that blood relatives of the search \textit{probe} are at or near top rank. Inherently, the paradigm of search \& retrieval mimics the use-case where family members (\ie parents and other relatives) for a missing child-- the missing child is the \emph{probe} and the relatives and ``distractors'' make-up the \emph{gallery}.



\begin{figure}[t!]
    \centering
    \includegraphics[width = .9\linewidth]{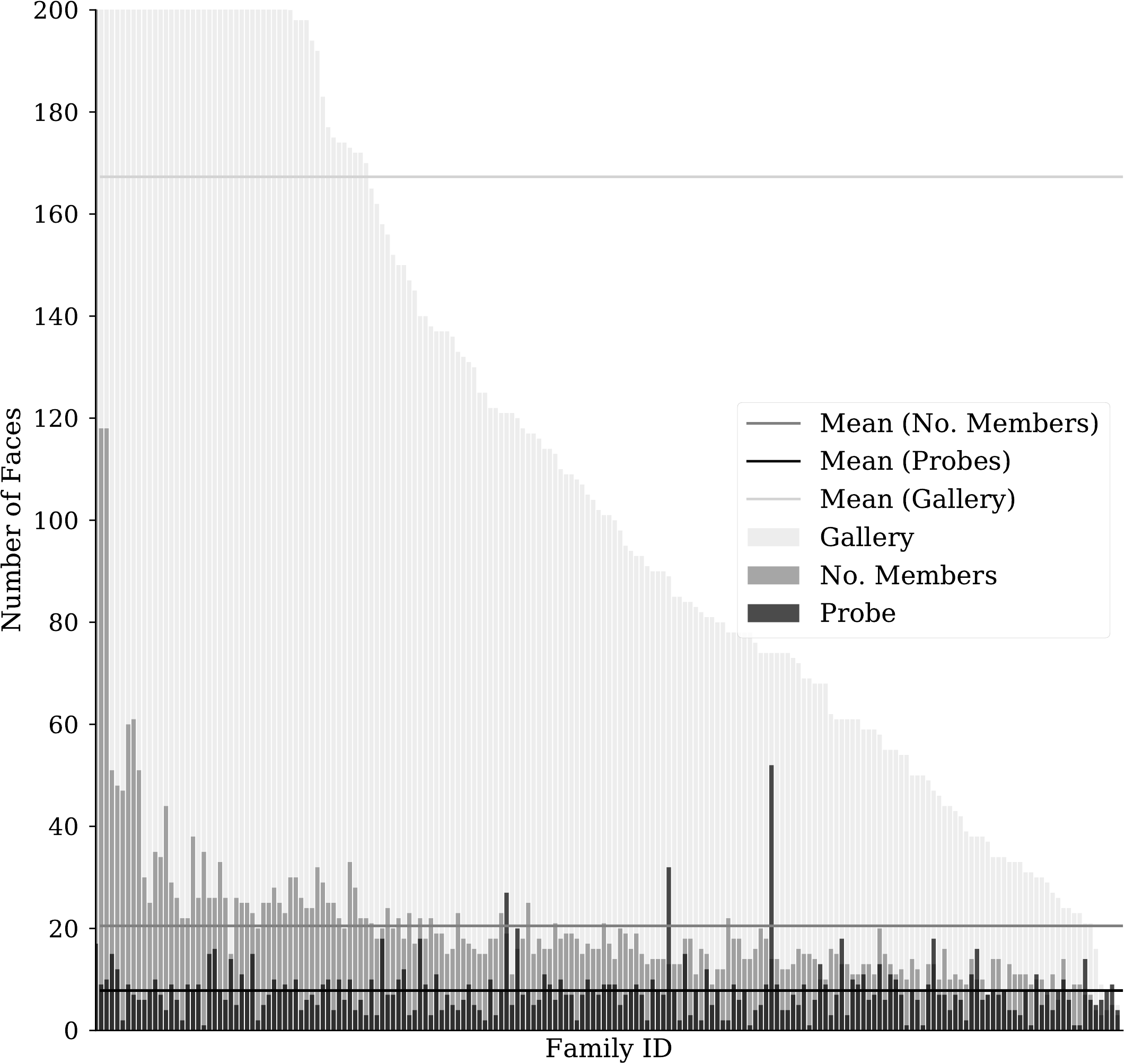}
    \caption{\textbf{Face counts per family in {\emph test} set of T3~\cite{robinson2020recognizing}}. The probes have about 8 faces on average, while the number of family members in the gallery nears 20 on average, with a total average of 170 faces.}
    \label{fig:track3:counts}
\end{figure}

\vspace{1mm}\noindent\textbf{Data splits.}
\textit{Probes} are supported by a variable number of samples. The \textit{gallery} is made-up of all relatives for each \textit{probe}. At test-time, the goal is to leverage all samples of a given \emph{probe} to best generate a ranked list of all members in the \emph{gallery}. As listed in Table~\ref{tbl:track3:counts}, the \emph{gallery} is made-up of 31,787 facial images from 190 families (Fig.~\ref{fig:track3:counts}): inputs are templates per subject (\ie \emph{probes}), and outputs are ranked lists of the entire \emph{gallery}. The \emph{gallery} remains static for the entire experiment-- a variable number of true relatives ranges from few to many; however, there are relatives present for every \emph{probe} making the setting closed form. As mentioned, \emph{probes} are organized as templates, meaning there are a variable number of samples per subject, leaving the method of fusion an open research question.  

\begin{table*}[ht!]
\centering
\caption {\textbf{T1 results.} Averaged verification accuracy scores of \ac{rfiw}.}
\label{tab:benchmark:track1}
\begin{tabular}{r|ccc|cccc|cccc|c}
  \textbf{Methods}& \textbf{BB} & \textbf{SS} & \textbf{SIBS} & \textbf{FD} & \textbf{FS} & \textbf{MD} & \textbf{MS} & \textbf{GFGD} & \textbf{GFGS} & \textbf{GMGD} & \textbf{GMGS}  & \textbf{Avg.} \\
  \midrule
  ArcFace~\cite{wang2018additive} (baseline)& 0.57 & 0.64 & 0.50 & 0.61 & 0.66 & 0.69 & 0.62 & 0.66 &0.71& 0.73 & \textbf{0.68}  & 0.64\\
    stefhoer~\cite{id2} & 0.66 & 0.65 & 0.76& \textbf{0.77} & 0.80 & 0.77 & \textbf{0.78} & 0.70 & \textbf{0.73} & 0.64 & 0.60  & 0.74\\
     ustc-nelslip~\cite{id6}& 0.75 & 0.74 & 0.72& 0.76 & 0.82 & 0.75 & 0.75 & \textbf{0.79} & 0.69 & \textbf{0.76} & 0.67  & 0.76\\
     DeepBlueAI~\cite{id3}& 0.77 & 0.77 & 0.75 & 0.74 & 0.81 & 0.75 & 0.74 & 0.72 & \textbf{0.73} & 0.67 & \textbf{0.68}  & 0.76\\
  vuvko~\cite{id4}& \textbf{0.80} & \textbf{0.80} & \textbf{0.77}& 0.75 & \textbf{0.81} & \textbf{0.78} & 0.74 & 0.78 & 0.69 & \textbf{0.76} & 0.60  & \textbf{0.78}\\
\end{tabular}
\end{table*}

\vspace{1mm}\noindent\textbf{Settings and metrics.} 
Each subject (\ie \textit{probe}) is searched independently-- 190 subjects with one-to-many faces. Hence, 190 families make up the \textit{test} set. Following template conventions of other \textit{many-to-many} face evaluations~\cite{whitelam2017iarpa}, face images of unique subjects are separated by identity. The \textit{gallery} then contains a variable number of relatives, each with a variable number of faces.

Mean average precision (MAP) was used to measure system performances. Mathematically speaking, average precision (AP) is first calculated per \emph{probe} as follows:
$$\text{AP}(f)=\frac{1}{P_F}\sum^{P_F}_{tp=1}Prec(tp)=\frac{1}{P_F}\sum^{P_F}_{tp=1}\frac{tp}{rank(tp)},$$
where AP is a function of family $f$ with a total of ${P_F}$ \ac{tpr}. Then, MAP is found by averaging over $N$ \emph{probes} via
$$\text{MAP} = \frac{1}{N}\sum^{N}_{f=1}\text{AP}(f).$$

\begin{table}[!b]
    \centering
    
    \caption{\textbf{T2 Counts}. No. pairs (\textbf{P}), families (\textbf{F}), face samples (\textbf{S}).}
    \begin{tabular}{p{.1in}m{.1in}ccc}
    & &\textbf{FMS} &\textbf{FMD} &\textbf{Total}\\\hline
     \parbox[t]{2mm}{
     \multirow{3}{*}{\rotatebox[origin=c]{90}{\textit{Train}}}}&\textbf{P} & 662  & 639 &1,331 \\
    \multirow{3}{*}{} &\textbf{F}  &375 & 364 & 739\\
    \multirow{3}{*}{} &\textbf{S} &8,575& 8,588 &  17,163\\\hline
    
    \parbox[t]{2mm}{
    \multirow{3}{*}{\rotatebox[origin=c]{90}{\textit{Val}}}} &\textbf{P}  & 202 & 177 & 379 \\
    \multirow{3}{*}{} &\textbf{F}  &116  & 117& 233\\
    \multirow{3}{*}{} &\textbf{S}  & 2,859 & 2,493 & 5,352 \\\hline
    \parbox[t]{2mm}{
    \multirow{3}{*}{\rotatebox[origin=c]{90}{\textit{Test}}}} &\textbf{P}  &  205& 178 & 383  \\
    \multirow{3}{*}{} &\textbf{F}  & 116  & 114  & 230 \\
    \multirow{3}{*}{} &\textbf{S}  & 2,805 &2,400 &5,205\\\hline
    
    \end{tabular}\label{tbl:track2:counts} 
\end{table}

\vspace{1mm}\noindent\textbf{Baseline and results.}
Submissions consisted of a matrix with a row per \emph{probe} listing the indices of all subjects in the \textit{test} gallery as a ranked list. Results are listed in Table \ref{tbl:t3:benchmarks} with sample inputs and predictions shown in Fig.~\ref{fig:track3:montage}.

%


\color{review2}
\section{Methodologies}\label{sec:methods}\color{review}
Many formulated kinship recognition problems in the visual domain as multi-view, multi-task, and multi-modal, which is typically to increase the amount of information obtainable, even when the final target is among other targets during training (\ie auxiliary tasks that complement the knowledge obtained from recognition, alone). For instance, the \ac{dkmr} was proposed as a jointly-trained model on top of a graph optimization algorithm~\cite{liang2017using}. Clearly, deep learning has overcome the traditional metric-learning approaches from about 2017~~\cite{li2017kinnet, qin2018heterogeneous, wei2019adversarial, zhu2019visual, mukherjee2019kinship} and still today~\cite{zhang2020advkin, laiadi2020tensor, wang2020kinship, multiperson2020, AsianConferenceonPatternRecognitionAucklandN}. \color{review2}We will first review the traditional methods, and then deep learning, for discriminating problems; an overview of the kin-based generative modeling is given at the end of the section.\color{review}

\begin{figure}[t!]
    \centering
    \includegraphics[width =.99\linewidth]{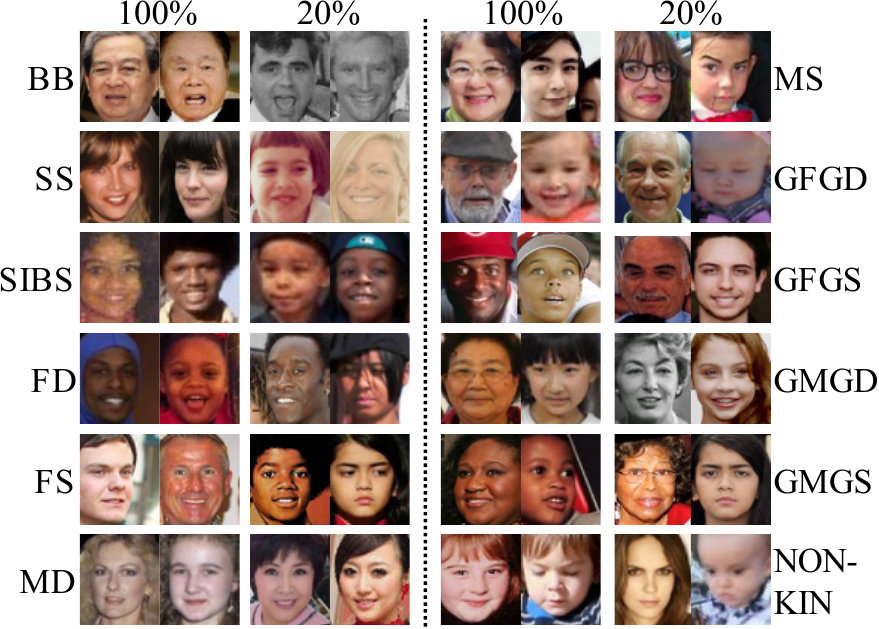}
    \caption{\textbf{Qualitative analysis of T1~\cite{robinson2020recognizing}.} Samples of each relationship type that all of the teams either got correct (100\%) or mostly not (20\%) for the elevin pair types of \ac{fiw} and \emph{NON-KIN}.}
    \label{fig:track1:samples:submitted}

\end{figure}

\color{review}
\subsection{Traditional approaches}

The main focus of the survey is on large data resources, along with the modern-day complex, data-driven modeling (\ie deep learning). However, such respective work makes up the latter half of the decade. Prior, feature and metric-learning dominated the first half. Furthermore, several works are more recent. For completeness, we will briefly discuss several methods that predate the deep methods on \ac{fiw} - results reported from the respective works will be for benchmark UBKin and KinWild datasets (Table~\ref{tab:kinwild}). 

\color{review2}
\vspace{1mm}\noindent\textbf{Handcrafted features}. Fang~\etal proposed using features such as geometric differences between face parts, color features, and handcrafted features that were the basis for the metrics to be learned in the years to come~\cite{fang2010towards}. Furthermore, and as mentioned, many of the smaller datasets are limited in diversity (\ie all similar demographics) and with pairs from the same photos, from which some proposed color-based features~\cite{crispim2020verifying}. Still, papers that hone-in on the smaller data employ more classical approaches, such as representation learning via binary trees~\cite{RAVIKUMAR2020e03751}.

\vspace{1mm}\noindent\textbf{Metric learning}. \color{review}Metric learning methods are popular solutions in kin-based vision problems. The general idea is to optimize a metric between classes. In kinship verification, the classes are \emph{KIN} and \emph{NON-KIN} (\ie true match and imposter, respectively). Lu~\etal proposed \ac{nrml} for kinship verification which aims to learn a distance metric that pulls image pairs without kin relations as far as possible and minimizes the distance between image pairs with kin relations~\cite{yan2014discriminative}. Yan~\etal extended this idea to discriminative multi-metric learning (DMML) method, which leveraged multiple types of features~\cite{yan2014prototype}. Kohli~\etal proposed a filtered contractive deep belief net (fcDBN) made by stacking fcRBMs to learn weights in a greedy, layer-by-layer fashion using both local and global features~\cite{kohli2016hierarchical}.

Wu~\etal combined color and texture features for kinship verification with extreme learning machines (ELM) for robustness on small data~\cite{wuX2018kinship}. Mahpod~\etal proposed a hybrid asymmetric distance learning (MHDL) scheme, combining symmetric and asymmetric multiview distances~\cite{mahpod2018kinship}. Most recently, Hu~\etal proposed treated different features as multiple views via a multi-view geometric mean metric learning (MvGMML)~\cite{hu2019multi}.

\color{review2}
For more details on the traditional methods, see~\cite{qin2019literature}. \color{review}

\begin{table}[!b]
\centering
\caption {\textbf{Verification scores.} Results for tri-subject (\ie T2).}
\label{tab:benchmark:track2}
\begin{tabular}{r|ccc}
  &\textbf{FMS} & \textbf{FMD} & \textbf{Avg.} \\
  \midrule
  
  Sphereface~\cite{wang2018additive} (baseline) & 0.68 & 0.68 & 0.68 \\ 
    stefhoer~\cite{id2} & 0.74 & 0.72 & 0.73 \\
  DeepBlueAI~\cite{id3}  & 0.77 & 0.76 & 0.77 \\
 ustc-nelslip~\cite{id6}  & \textbf{0.80} & \textbf{0.78} & \textbf{0.79} \\
\end{tabular}
\end{table}

\color{black}

\subsection{Deep learning approaches}
\color{review}
The 2012 AlexNet~\cite{krizhevsky2012imagenet} sparked the deep learning era. As done in many problems, deep learning grew more popular with the big data provided with \ac{fiw}. Still on \ac{kfw}, we first review the deep metric down using small amounts of training data, and then discuss the data-driven work done using \ac{fiw}.


There are many commonalities between the different solutions proposed as part of the \ac{rfiw} challenge. Typically, a ResNet-based~\cite{he2016deep} backbone; if not, then together with FaceNet~\cite{schroff2015facenet}. Nonetheless, the story as seen in the timeline is split in half (\ie with the latter half dominated by modern-day deep learning approaches) and quite significantly, metrics learned on top of hand-crafted features dominated the charts as \ac{sota} for many years~\cite{8803754}. As recent as 2017, metric-learning was a go-to approach for kin-based problems, whether a single metric or multiple (\eg \ac{lm3l}~\cite{hu2017local}). Even so, geometric and distant features in pixel space (\eg key point coordinates on neutral face~\cite{KALJAHI2019100008})-- directly related to insufficient data for modern-day data-driven machinery (\ie deep learning). 

Provided a deep \ac{cnn} trained to classify face identity, the encodings produced encapsulate much information of the subject. However, instead of looking for absolute closeness in embedding space as the ideal case for a set of samples of a single class (\ie identity), in kin-based tasks we hope to detect when similarities between a pair (or group) of faces (\ie encoded) reflect that of the various relationship-types. For this, many tend to fine-tune models initially trained on a larger \ac{fr}-based database, such as
VGG-Face~\cite{schroff2015facenet}, VGG2~\cite{cao2018vggface2}, and MSCeleb~\cite{guo2016ms}. \color{review2}We next speak on various flavors of deep learning.

\begin{figure}[t!]
    \centering
    \includegraphics[width =.9\linewidth]{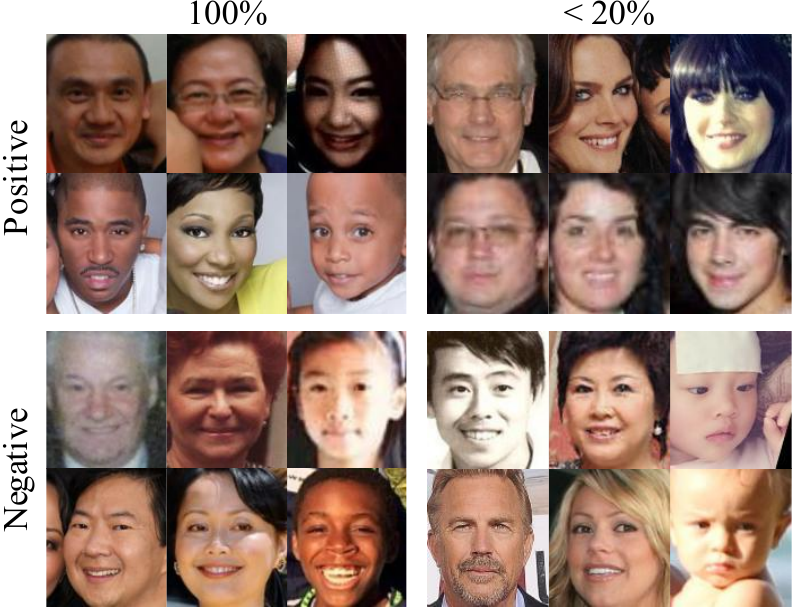}
    \caption{\textbf{Sample of T2~\cite{robinson2020recognizing}.} Samples that all teams got correct (left) and mostly incorrect (right) for FMS (top rows) and FMD (bottom).}
    \label{fig:track2:samples:submitted}
\end{figure}

\begin{figure}[t!]
    \centering
    \includegraphics[height=1.65in,angle=-90,origin=c]{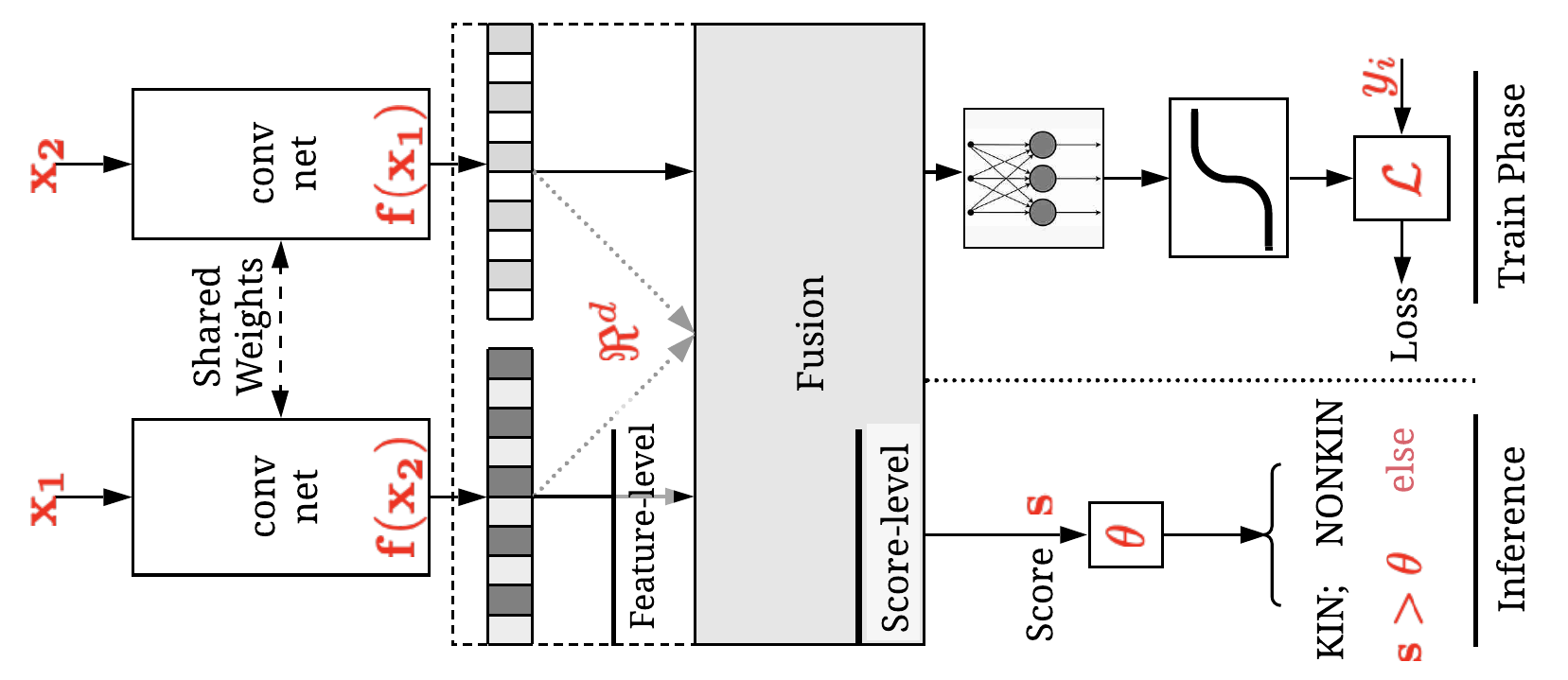}
    \caption{\textbf{Generic Siamese network.} Approaches tend to follow the Siamese model, differing in method of fusion, \ie \emph{black-box}, middle. Specifically (from \emph{top}-\emph{to}-\emph{bottom}), an image pair shot $x_1$ and $x_2$.}\label{fig:siamese}
\end{figure}

\vspace{1mm}\color{review2}
\noindent\textbf{Pre-trained CNNs.} \color{black}Besides, most solutions involve the renowned Siamese training model, and many of which still incorporate a cosine loss as in the seminal work done at Bell Lab's mid-90s~\cite{bromley1994signature}, \ie multiple inputs to networks with shared weights for which metric is learned on top (Fig~\ref{fig:siamese}). In the simplest form, Siamese-based \ac{cnn} models map two or more samples by a single \ac{cnn} to a real-number vector space $\mathrm{R}^d$ (\ie a function $f(\cdot)$ to encode an image (\ie facial [encoding, embedding, feature] of size $d$, especially in the context of facial representation, all refer to the $f(x_i)=z_i\in\mathrm{R}^d$. Generally, and in most methods proposed in \ac{rfiw}, the shared model is pre-trained data for another, yet similar task (\ie facial recognition). With that, the \ac{cnn} that now serves as an encoder, maps $k$ samples to its $d$-dimensional space learned to discriminate between faces. With the Siamese frozen-- whether entire network, with a couple of layers on top set with a small learning rate, or popped off by adding a path that splits off prior to later rejoin or just remove entirely-- the goal then is to learn a metric optimal for recognizing family members by face cues. Clearly, there are several design choices-- with simple solutions in those with an \emph{off-the-shelf} \ac{cnn} with no additional training (\ie trained for \ac{fr}, so naively assuming that the best way to detect kinship is to detect faces that look like the source). However simple, and with many cases a fair assumption, the naive approach outperformed previous \ac{sota} methods prior to \ac{fiw} providing the number of data samples needed to suffice the capacity of most deep learning approaches. In light of this, the \ac{cnn} then serves as the method for feature extraction-- claiming to provide the best face representations for the task. As previously described of the wave of metric-based and subspace-modeling methods, we can then further refine the output of the feature extractor by extending the composition function by adding and training mappings in the embedding space and while again, often with the weights of the pre-trained \ac{cnn} $f$ held static. From this, kin-based tasks can be targeted by learning filters, mappings, and even metrics from the embedding space on up (\ie build up from the embedding space from where face embeddings are compared in some fashion).

\begin{table}[!b]
    \centering
    \caption{\textbf{T3 Counts.} Individuals \textbf{I}, families \textbf{F}, face samples \textbf{S}.}
    \begin{tabular}{p{.1in}m{.1in}ccc}
    & &\textbf{Probe} &\textbf{Gallery} &\textbf{Total}\\\hline
     \parbox[t]{2mm}{
     \multirow{3}{*}{\rotatebox[origin=c]{90}{\textit{Train}}}}&\textbf{I} & --  & 3,021 & 3,021 \\
    \multirow{3}{*}{} &\textbf{F}  &-- & 571 & 571\\
    \multirow{3}{*}{} &\textbf{S} & --& 15,845 & 15,845 \\\hline
    
    \parbox[t]{2mm}{
    \multirow{3}{*}{\rotatebox[origin=c]{90}{\textit{Val}}}} &\textbf{I}  & 192 & 802  & 994  \\
    \multirow{3}{*}{} &\textbf{F} & 192 & 192 & 192  \\
    \multirow{3}{*}{} &\textbf{S}  &1,086  &4,030 &5,116 \\\hline

    \parbox[t]{2mm}{
    \multirow{3}{*}{\rotatebox[origin=c]{90}{\textit{Test}}}} &\textbf{I}& 190 & 783  & 9d73 \\
    \multirow{3}{*}{} &\textbf{F} &190  & 190  & 190   \\
    \multirow{3}{*}{} &\textbf{S}  &1,487  & 31,787 & 33,274\\\hline
    
    \end{tabular}\label{tbl:track3:counts} 
\end{table}

\vspace{1mm}\color{review2}
\noindent\textbf{Deep metric learning.} 
\color{black}Lu~\etal proposed to learn a distance metric for $K$ feature types via $K$ MLPs - learn to project each feature using the optimal thresholds determined independently~\cite{lu2017discriminative}. This method, which was called discriminative deep metric learning (DMML), proved effective on the \ac{kfw} settings of minimal training data (Table~\ref{tab:kinwild}). 

\vspace{1mm}\color{review2}
\noindent\textbf{Fine-tuning.} 
\color{black}There is an abundant of public \ac{fr} data (\eg LFW, VGG, MSCeleb~\cite{guo2016ms}) with some labeled by \emph{soft attributes} (\eg age~\cite{zhifei2017cvpr}, gender~\cite{cheng2019exploiting}, attribute, and diverse demographics~\cite{robinson2020face, DBLP:journals/corr/abs-1812-00194}). With this, and provided the known concept of deep learning tending to learn transferable features~\cite{zhang2018survey}, the use of fine-tuning pre-trained has been done by many. For instance, a SphereFace loss, which is a multi-class loss, is first used to train a large \ac{cnn} to do facial recognition on identities of an auxiliary dataset, and then having the layers near the top fine-tuned to recognize the families of the \ac{fiw} training set via

\begin{equation}
    \mathcal{L}_{\text{family}}(\theta) = -\frac{1}{B} \sum_{i=1}^{B}\log\frac{\exp^{W_{y_i}^Tx_i+b_{y_i}}}{\sum_{j=1}^{N}\exp^{W_{y_i}^Tx_i+b_{j}}},
    \label{eq:arcfamily}
\end{equation}
where $B$ is the batch size, $N$ is the number of families, $x_i$ is the face encoding from family $y_i$, $W$ is the weight matrix (\ie $W_j$ denotes the $j^{th}$ column) and $b$ is the bias term. In the end, verifying kinship between a face pair can be done using the model to encode the faces and cosine distance to measure their closeness. If family labels are unavailable, which is another setting of the verification task, approaches tend to use Siamese concepts on top of the pre-trained \ac{cnn} (Fig.~\ref{fig:siamese}). Specifically, sharing weights for two or more samples, and penalizing based on the closeness between a set of samples upon being encoded by the network, has shown to be an effective means of staging a network for the verification task. In return, Siamese; furthermore, the relationship between the pairs with respect to labels at training differences is in preprocessing, method of fusion (\eg \emph{early} versus \emph{late}).
\begin{figure}[t!]
    \centering
    \includegraphics[width = .94\linewidth]{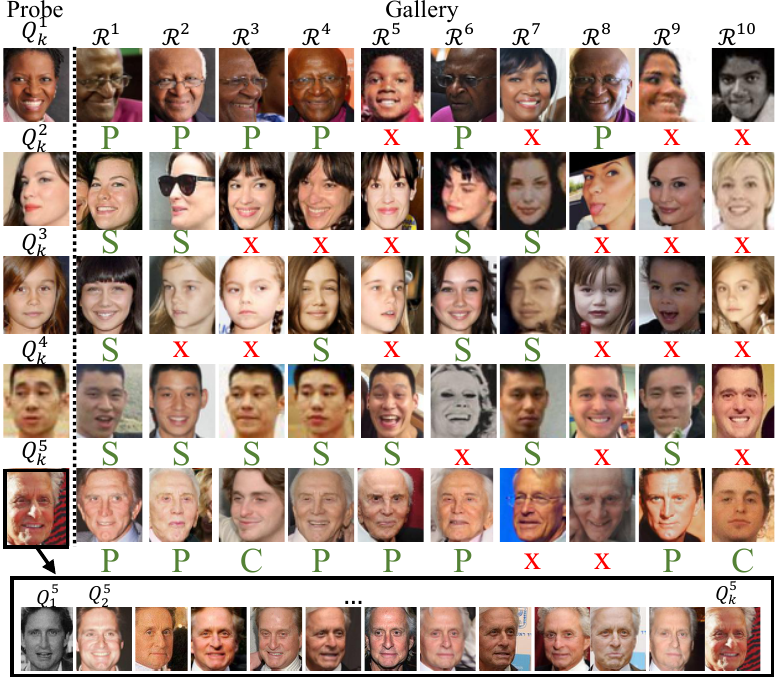}
    \caption{\textbf{T3 sample results~\cite{robinson2020recognizing}.} Each query (row) for a given \emph{probe} (left column) had all samples in the gallery returned in a ranked list - here we show top 10. \ac{fp} are labeled by \textcolor{red}{x}, while true matches list the relationship type as \textcolor{ao(english)}{P} for parent, \textcolor{ao(english)}{C} for child, and \textcolor{ao(english)}{S} for sibling.}
    \label{fig:track3:montage}
    \vspace{-1mm}
\end{figure}

{\color{review}

\begin{table}[b!]
	\centering

	\caption{\textbf{T3 results.} Performance ratings for \ac{sota} methods.}

	\begin{tabular}{r|cc} 
	      \textbf{Methods}  &\textbf{mAP} & \textbf{Rank@5} 	\tabularnewline \hline
		  Baseline (Sphereface)~\cite{wang2018additive} & 0.02 & 0.10	\tabularnewline
		  
		  DeepBlueAI~\cite{id3} & 0.06 & 0.32	\tabularnewline

		  HCMUS notweeb~\cite{id9} & 0.07 & 0.28		\tabularnewline
		  ustc-nelslip~\cite{id8} & 0.08 & 0.38		\tabularnewline
		  vuvko~\cite{id4} & \textbf{0.18} & \textbf{0.60}		\tabularnewline
	\end{tabular}
	\label{tbl:t3:benchmarks}
\end{table}

}
In~\cite{id8}, Track I and III completed in succession, such that a wider sweep of \ac{cnn} backbones, loss functions, and fusion methods were assessed in Track 1, to both gain deeper understanding to make decisions pertaining to Track III. Mainly, ResNet50 and SENet50 were evaluated separately, each with additional fully-connected layers with two losses on top, \ac{bce} and Focal loss. \ac{bce}, a widely used loss that does as its name implies: uses the measure of entropy of a distribution, say $q(y)$ for $c\in{1, \dots, C}$ classes as $\mathrm{H}(q) = \sum_{c=1}^Cq(y_c)*\log(q(y_c))$. Since we have no knowledge of the true distribution, we aim to match samples of the \emph{true} distribution $p(y)$. Hence, cross-entropy is entropy between $p(y)$ and $q(y)$.

Yu~\etal found that \ac{bce} loss outperformed Focal Loss for all fusion schemes and settings in Track I~\cite{id8}. Intuitively, this makes sense as Track I, a Boolean task, has an equal number of positive and negative pairs-- imbalanced data motivated Focal Loss, which is not an issue for verification. Then, transferring over the model, loss, and fusion settings that worked best for Track I to Track III and used as is. The difference is in the ranking scheme (\ie provided multiple faces per query, the average of all faces and each gallery sample determined the score at the subject-level.

\vspace{1mm}
\noindent\textbf{Deep representation.} 
\color{review}
Training a set of \acs{cnn}, each targeting specific regions (or parts) of the face, was proposed as a solution for \ac{kfw}~\cite{zhang122015kinship}.
Then, 
\color{black}\ac{hsl} tackled various tasks of kinship recognition via multi-view learning, with the different views set as different relationship types dubbed~\ac{msml}.~\cite{qin2018heterogeneous}. Similarity, \ac{svdd} was proposed as a \ac{sml} loss function, allowing detailed information to be extracted as geometric and appearance-based features for kinship verification~\cite{qin2019social}. 
Duan~\etal proposed a coarse-to-fine scheme for which \acs{cnn} at different levels (\ie layers) were transferred from being trained using a \ac{fr} dataset and then fine-tuned for kinship using a loss function based on \ac{nrml}~\cite{duan2017face}. In fact, many recent works leveraged existing \ac{fr} methodologies (\eg \ac{cnn} trained to classify faces) as a prior, the then fine-tune using the kin-based image data as the source in a transfer-learning regime~\cite{zhang2019deep}.

\begin{figure}[!t]
    \centering
    \includegraphics[width=\linewidth]{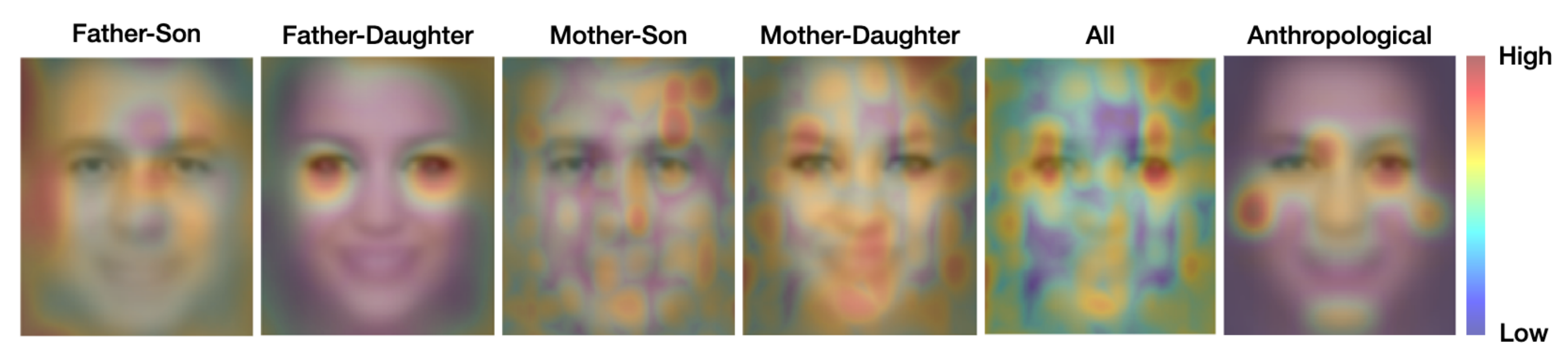}
    \caption{\textbf{Activations from mapping image-to-latent space (from~\cite{dehghan2014look}).} The salience mapped from the activation response and superimposed on the average face. Family101 dataset was used for this experiment~\cite{fang2013kinship}. The end result depicted here were dubbed the \emph{genetic features} from latent space of a trained Gated \ac{ae}.}
    \label{fig:genetic:features}
\end{figure}

Several lines of research specifically focused on the {one-to-one} kinship verification problem by learning a face encoder robust in detecting kinship relationship \color{review}via \ac{ae} (\eg \ac{dae}~\cite{wang2018cross, chergui2020kinship, chukinship}), \ie deep representation learning methods~\cite{kohli2018deep}.  Dehghan~\etal was amongst the first, proposing to train a Gated \ac{ae} to encode faces as \emph{genetic features}, and weighting according to the salience for the respective relationship type~\cite{dehghan2014look}. Fig.~\ref{fig:genetic:features} depicts the salience, with high being most similar regions and low dissimilar.
Besides still-faces, deep learning approaches were also proposed for recognizing kinship pairs using facial cues in video data~\cite{sun2018video}. A sequence recurrent \ac{nn} was trained for kinship verification in videos using a novel attention mechanism~\cite{lv2019attentive}. With videos come more bits of information; however, the range of bits (\ie the underlying variation of the data) should be optimized to maximize the information gain. In other words, video data introduces another space for fusion in the choosing of the best frame(s) to describe and represent~\cite{gong2019video}. Graphical neural network (GNN) with a metric learned on top proved to be one of the most effective deep learning models employed for kin-based vision problems~\cite{liang2018weighted}. \color{review}Not just on the large-scale \ac{fiw} data, but a graph-based kinship reasoning (GKR) network proved effective on \ac{kfw}~\cite{li2020graph} (Table~\ref{tab:kinwild}).\color{black}

\noindent\textbf{Approaches to data challenges.}
\ac{rfiw} serves as a platform for experts to publish and junior scholars to get started. The first edition of \ac{rfiw} was in 2017~\cite{robinson2017recognizing} - a data challenge workshop in conjunction with the \emph{ACM Conference on Multimedia}. Ever since, \ac{rfiw} has been held annually (\ie 2018-2020 held in conjunction with \ac{fg} as a data challenge), with each year building on the prior. Let us review series highlights over the years, and then focus on the top teams of the 2020 edition.


From the start, solutions for \ac{rfiw} typically involved \ac{cnn}s pre-trained for \ac{fr}. 
For the top performing submission of the 2017 \ac{rfiw}, Yong~\etal used an ensemble of deep \acs{cnn}s with data augmentation and mining techniques called KinNet~\cite{li2017kinnet}. Specifically, the authors proposed to train four ResNet models (\ie 80, 101, 152, and 269 layers) for \ac{fr} to then fine-tune for kinship verification via a triplet loss targeting intra-family relationships. KinNet used two tricks during training: (1) augmentation using imaging processing techniques (\eg gamma correction, down/up sampling of pixels, blurring) and hard-negative mining for selecting triplets. In the end, KinNet scored an impressive average of 74.9\%. It is important to note that the data has changed since this first edition of RFIW (\eg \emph{grandparent}-\emph{grandchild} types were not included). Thus, a comparison with the proceeding years would be unfair. 

In 2018, Dahan~\etal got the top performance 68.2\%~\cite{dahan2018selfkin}. Specifically, the authors trained a VGG-Face model with the novel \emph{local features conv-layer} that fused the Siamese inputs by summing the features. In other words, conventional conv-layers share weights in image space, whereas these authors proposed learning local weights to produce pair and location specific features. 

\color{review}In 2019, Laiadi~\etal extended XQDA-to-TXQDA to operate on multilinear data in a low dimensional and discriminative tensor subspace. TXQDA uses multilinear projections of tensors to a space with greater separation between data classes is enhanced in a way that helps lightening the small sample size problem (\ie results for both \ac{kfw} and \ac{fiw})~\cite{laiadi2019kinship}. Nandy~\etal, on the other hand, followed a Siamese learning approach~\cite{nandy2019kinship}, which we will next learn was the most common in 2020.\color{black}

\begin{figure}[t!]
    \centering
    \includegraphics[width=.95\linewidth]{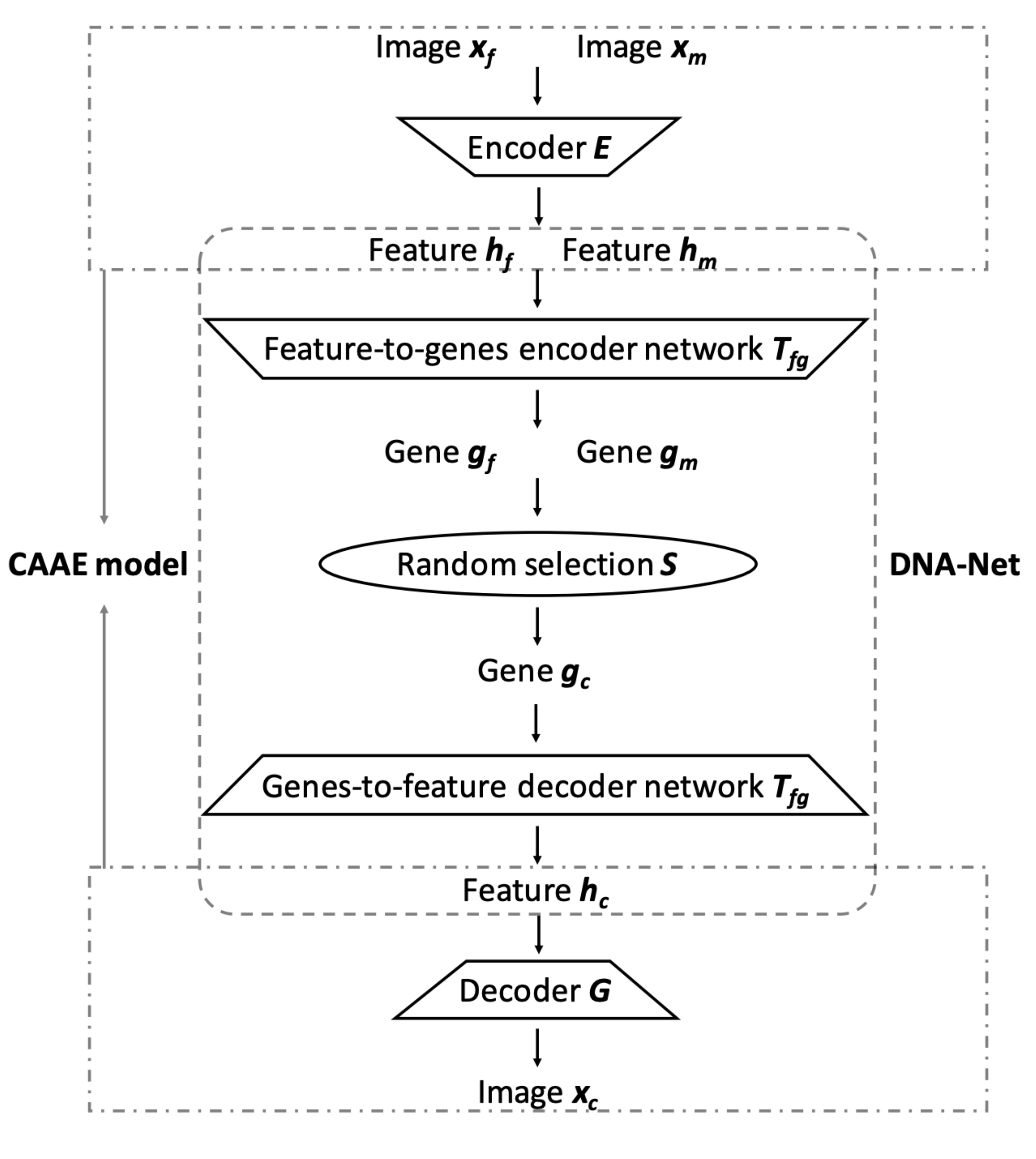}
    \caption{\textbf{Model to synthesize children faces from a parent-pair (visualizations from~\cite{gao2019will}).} Notice that the output of encoder $E$ is the concatenation of features from prospective parents, the father $h_f$ and mother $h_m$ joined by $\oplus$ such that the two embeddings encoded by the Siamese network are fused (\ie $2*\mathbb{R}^d\rightarrow\mathbb{R}^{2d}$) before passed as input to the \ac{caae} model.}\label{fig:dna:net}
\end{figure}




\color{review}The 2020 edition saw a great increase in interest and participation. Several methods were used as solutions for two or more tasks. 
\color{black}Shadrikov~\etal treated the different pair types as multiple tasks, training a local expert for each on top of a ResNet50, simultaneously. The authors used T1 data as validation, but then deployed on T2 and T3 as well. \color{review}Another multi-task approach applied \color{black}different fusion techniques in deep feature space~\cite{id6, id8}. 
\color{review}Zhipeng~\etal used two pre-trained \acp{cnn}, fused the two face encodings by different types of arithmetic, and generated solutions for all three tasks~\cite{id6, id8}. In \cite{id3}, the distance between faces was then determined euclidean distance (instead of the typical cosine similarity); also, SENet~\cite{iandola2016squeezenet} was used as the backbone showing a modest boost over ResNet50 on the validation, but dropping on the test. Like in~\cite{robinson2018visual}, except the authors now used Arcface instead of Sphereface, \"{H}\"{o}orman~\etal fine-tuned a \ac{cnn} using families as the classes\cite{id3} - ultimately placing second in verification (\ie T1) and tri-subject (\ie T2). \color{black}Yu~\etal put emphasis on the dependence of family identification accuracy for cross-gender versus same-gender pairs types~\cite{id2}. These researchers constructed a Kinship \emph{comparator} module that consisted of eleven ``local expert networks'' connected in series-- eleven networks corresponding to the eleven relationship types of T1. In the end, Stefhoer scored the best for parent-child types FD and MS in T1. Yu~\etal encoded features from face images via a Siamese \ac{cnn} with shared weights~\cite{id6}. ResNet50 or SENet50 was used as the backbone, both pre-trained on VGGFace2~\cite{cao2018vggface2}. Team ustc-nelslip also employed two loss functions, binary cross-entropy and focal loss, and fused the features using two algebraic formulae leading to \( 2 \times 2 \times 2 = 8 \) independent ``models.'' A unique feature was the construction of a ``jury system'' to combine outputs of different models to improve accuracy. With \cite{id4} the top-scorer in T2. Nguyen~\etal competed in T1 and T3\cite{id9}-- the authors use a Siamese CNN with FaceNet (Inception-ResNet-v1 trained with triplet loss) and VGG-Face (Resnet-50) pre-trained. The authors also implement ArcFace~\cite{wang2018additive} - a family of loss functions based on the geodesic distance between feature vectors which aim to discriminate the latent representation of deep \acs{nn}. \color{review}Samples that were unanimously classified correct or most incorrect are in Fig.~\ref{fig:track1:samples:submitted},~\ref{fig:track2:samples:submitted}, and~\ref{fig:track3:montage}, along with average performance ratings in Table~\ref{tab:benchmark:track1},~\ref{tab:benchmark:track2}, and~\ref{tbl:t3:benchmarks} for T1, T2, and T3, respectfully. 

\color{black}

\noindent\textbf{Summary.}
\color{black}Of all the methods, there is a common factor: the larger the age gap the higher percentage of \ac{fp} during evaluation. As mentioned, this was addressed early on with UB Face~\cite{shao2011genealogical} and, although fundamental to the analysis of results over the years, proposed models tended to acknowledge this as a challenge, but with no added mechanism to make robust to age-variations between \emph{parent}-\emph{children}. That is, until Wang~\etal proposed using \ac{gan} technology to synthesize younger versions of an input face. Specifically, and a clearly effective data augmentation approach, the authors trained generators for both genders to account for this while training a deep \ac{cnn} with a maximum margin loss to do boolean classification (\ie \emph{KIN} / \emph{NON-KIN}). As formalized in their work, domain $A$, for \emph{aged}, was the source and domain $Y$m, for \emph{young}, was the target. Provided paired data, the \emph{parent} aimed to transform $x_i\in A\rightarrow x_j\in Y$ with data distribution $x\sim p_A\rightarrow x\sim p_Y$. Having noticed that \ac{fiw}, which most closely matches real-world data, does not necessarily have parents at older ages (\ie \emph{aged}). Thus, the inputs could very well be parents as juveniles, or even during infancy. To mitigate the problem, for there are no age-labels provided in \ac{fiw}, focus was directed to constrain the output such to influence younger aged faces less so, than if faced with an elderly parent.

\begin{figure}[t!]
    \begin{subfigure}[t]{0.5\textwidth}
        \centering
        \includegraphics[width=.8\textwidth]{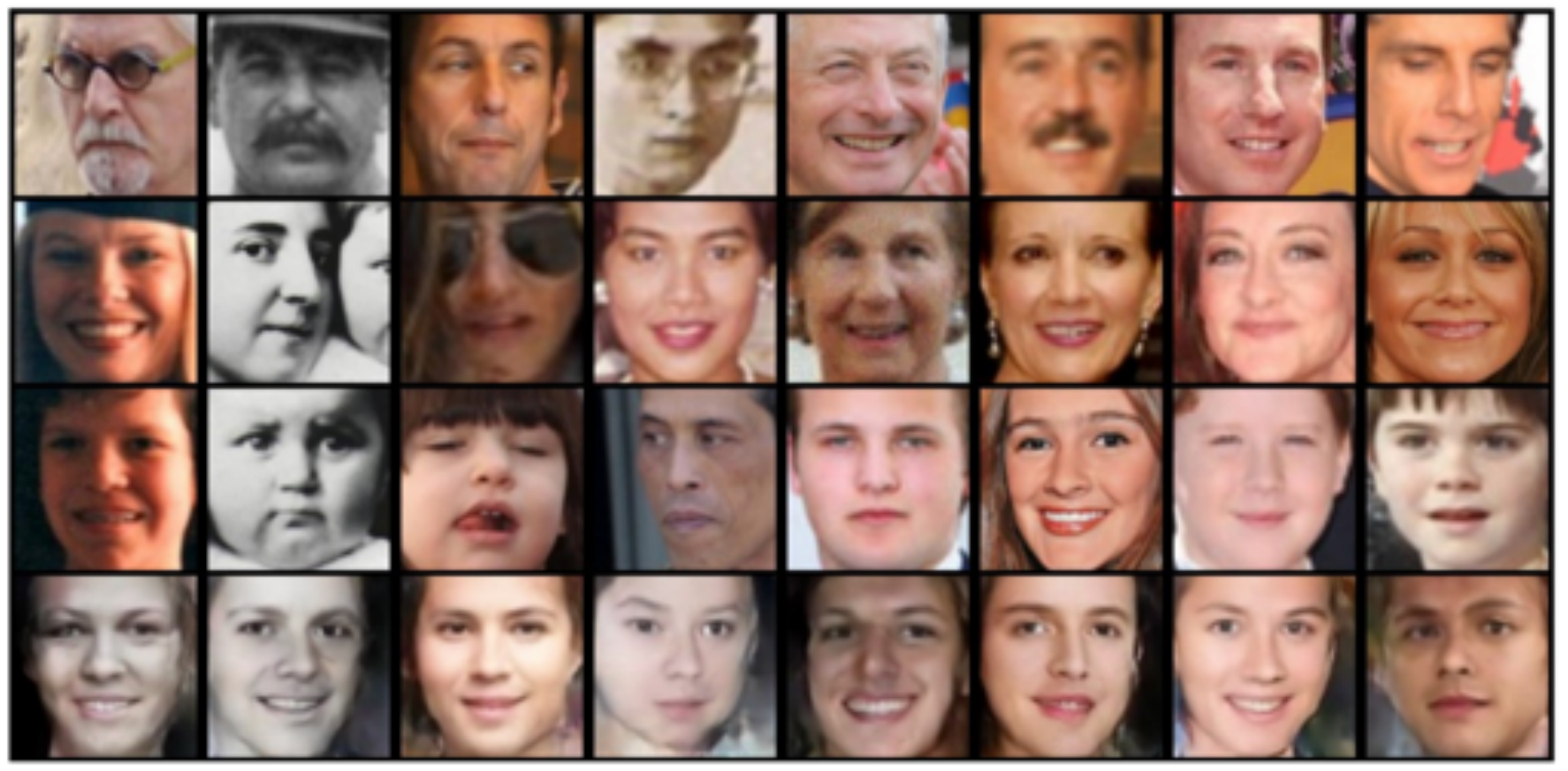}
        \caption{Random synthesis.}
        \label{fig:montage:random}
    \end{subfigure}%
    \\
    \begin{subfigure}[t]{0.5\textwidth}
        \centering
        \includegraphics[width=.8\textwidth]{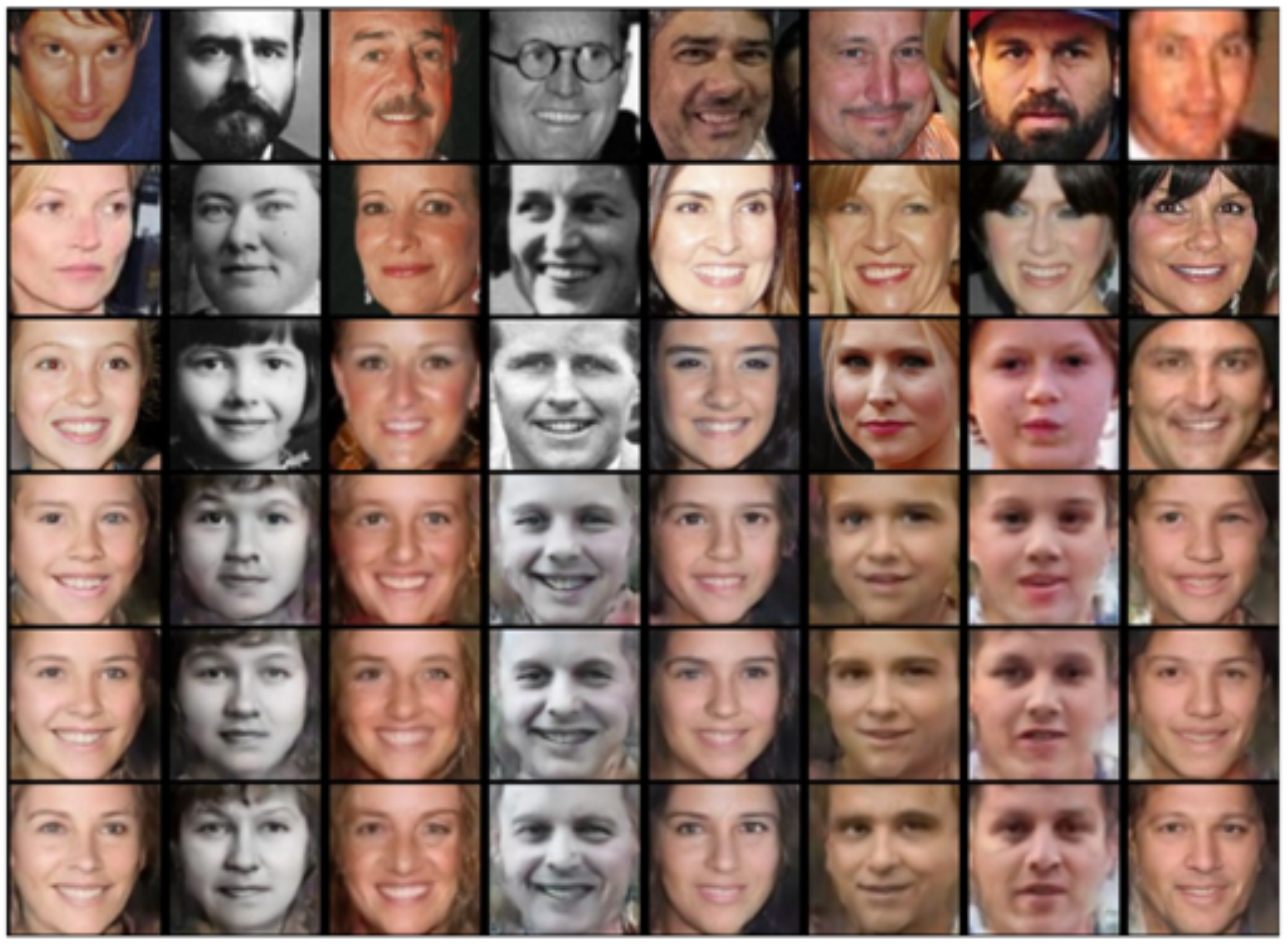}
        \caption{Age (\ie 10, 20, 30 years old from row 4-6, respectfully).}
        \label{fig:montage:age}
    \end{subfigure}%
    \\
    \begin{subfigure}[t]{0.5\textwidth}
        \centering
        \includegraphics[width=.8\textwidth]{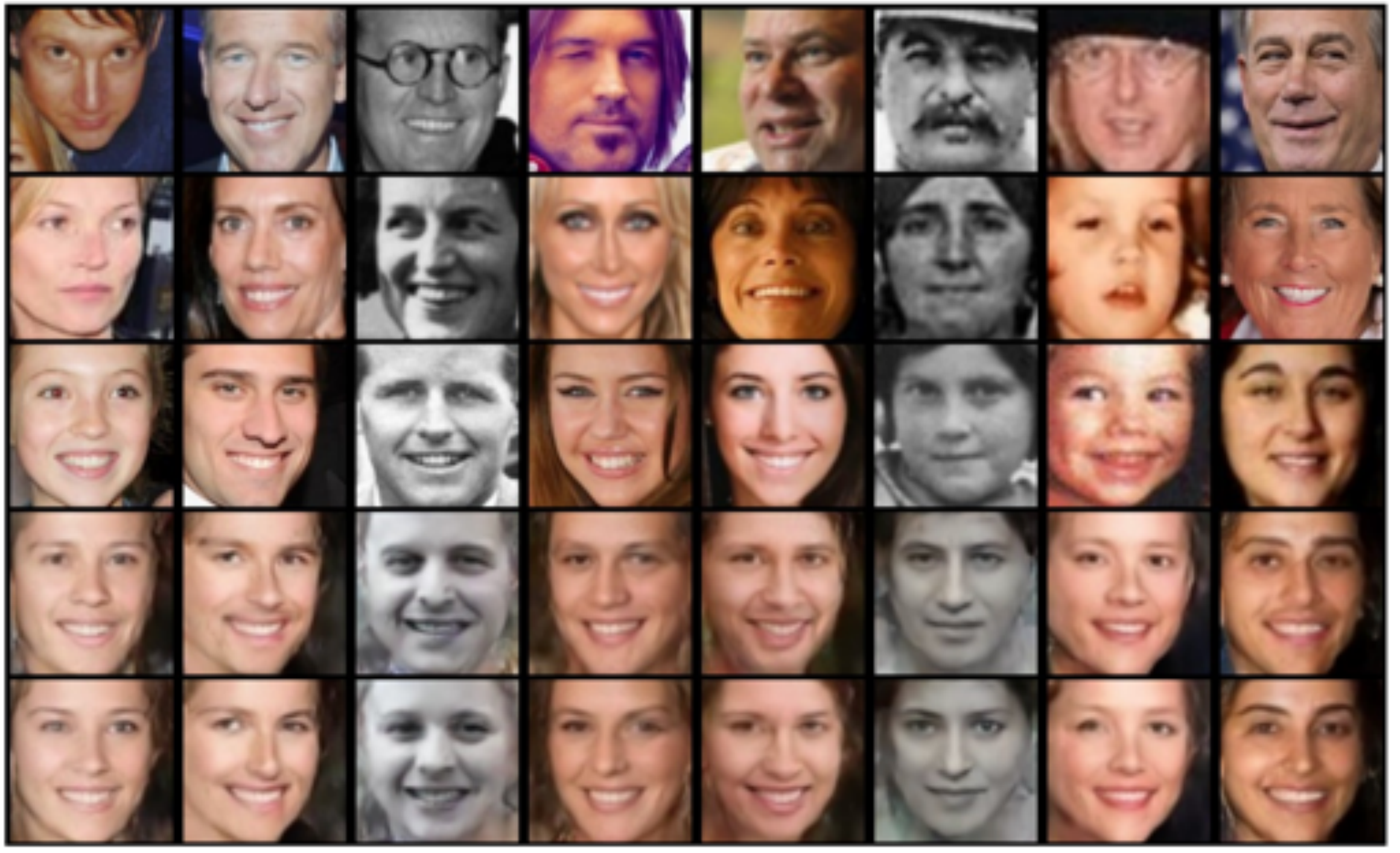}
        \caption{Gender (\ie male-to-female from row 4-5, respectfully).}
         \label{fig:montage:gender}
    \end{subfigure}
    \caption{\textbf{Synthesized results from~\cite{gao2019will}.} Columns correspond to families, with fathers on first row, mothers on second, and real and generated children on third row and bottom, respectively (a). See subcaption for specifics on age (b) and gender (c).}
    \label{fig:dna:net:sample:results}
\end{figure}

%% file: sections/applications.tex
\section{Applications}\label{sec:applications}
We next review the use-cases for kinship recognition technology.

\vspace{1mm}
\noindent\textbf{Entertainment and personal knowledge.}
 AncestryDNA claimed $>$15 billion people in its DNA network: their $>$3M paying subscribers (and $>$16M people DNA tested), resulted in the establishment of 100M family trees that form 13B connections across 80 countries.\footnote{\href{https://www.ancestry.com/corporate/about-ancestry/company-facts}{www.ancestry.com/corporate/about-ancestry/company-facts}} As of 2019, Ancestry launched AncestryHealth as a means to infer inheritable health conditions via DNA. Clearly, there is high interest in learning about one's family roots-- which started from curiosity (\ie knowing where one fits, recalling the aforementioned words of Furstenberg~\cite{furstenberg2020kinship}), but now includes learning about one's health from their DNA. Acquiring sufficient data to support both DNA and imagery would be difficult, at best; however, provided more reliable kinship recognition capabilities, such technology would certainly enhance popular services such as those provided by billion dollar companies (\eg \href{https://www.ancestry.com/}{ancestry.com}).

\vspace{1mm}
\color{review}\noindent\textbf{Connect families.}
\color{black}Identify unknown children being exploited online; reconnect families separated by the modern-day refugee crisis~\cite{mcnatt2018impact}; find unknown relatives, whether directly or indirectly. Statistics show that people want to learn of missing family ties. Furthermore, unfortunate scenarios leave family members desperate to reconnect with lost member(s). \color{review}Alternatively, law-enforcement could use kinship to solve other high-profile crimes-- the decades long mystery of who the \emph{Golden State Killer} was got solved by using DNA to build his family tree~\cite{goldenstate}. \color{black}

\color{review}
\vspace{1mm}
\noindent\textbf{Soft attribute as prior knowledge for traditional FR.}
\color{black}
\diff{Per Kohli~\emph{et al.}, using kinship as \emph{soft} information boosts the performance of face verification systems~\cite{kohli2018deep}. Hence,} whether it be to enhance \ac{fr} capabilities (\eg \diff{like done for attributes in}~\cite{taherkhani2018deep}), to learn to discriminate between hard negatives (\eg brothers), or to narrow the search (\eg \ac{fr} failed to identify bombers of the 2013 Boston Marathon) - but had we known they were brothers, the search space could have been drastically reduced. Hence, kinship provides a powerful cue to help boost existing \ac{fr} systems.

\vspace{1mm}
\noindent\textbf{Nature-based studies.}
With the new millennium came the ability of 3D scans of facial appearances of ten pairs of twins to be compared via landmark features (\ie anteroposterior and vertical facial parameters)~\cite{naini2004three}. About ten years later, this inspired Dehghan~\etal to ask: \emph{Who do I look like?} And then attempt to solve the question using computer vision (\ie gated \ac{ae}~\cite{dehghan2014look}).

\vspace{1mm}
\noindent\textbf{Kin-based face synthesis.} 
An early attempt to predict the appearance of a child from prospective parents was in~\cite{frowd2008predict}. Specifically, Frowd~\etal proposed EvoFit, which used classic shape-based modeling and \emph{eigenfaces} to project a pair of faces via statistical appearance-based modeling. In all fairness, the generative task was heavily influenced by~\cite{cootes2001active}, as many face synthesis tasks were throughout the years, and especially in 2006 the EvoFit came out. In short, EvoFit learned its weights from face samples collected in a tightly controlled setting-- per the requirement that 223 landmarks were precisely marked for all faces. As seminar as EvoFit was in its own right, this early attempt to predict the appearance of children was seemingly ahead of its time, in available machinery (lacking the data-driven, highly complex modeling techniques of today), in resources available to reproducible (\ie no public data released with paper), and in the problem statement itself. In other words, considering EvoFit was proposed before our 2010 timeline means it predated the first benchmark in kinship verification. With that, we believe the small impact of this work was due to its timing and, in return, the lack of complete support for the problem, so if others did want to partake they too would have to collect data. Meaning, it was impossible to reproduce results directly. Regardless how minimal the impact was in citations and usage of other researchers, the work certainly showed promise considering the results were from a minimally-sized data pool. Thus, had a widely used benchmark been practiced, or provided the data constraints were handled (\ie inability to generalize + inability for others to reproduce), then EvoFit could have attracted much more attention. Perhaps, our 2010 time-line would have had to start a few years prior. Nonetheless, this is only speculation and, therefore, we can only hypothesize the \emph{what ifs} after the fact.

%% file: sections/discussion.tex


\section{Discussion}\label{sec:discussion}
\subsection{Broader impacts}
The most recent \ac{rfiw} data challenge (\ie the 4th edition) seemed to attract far more attention than in the previous years-- task
T1 (verification) was repeated, and alongside more complex and practically motivated tasks T2 (tri-subject) and T3 (search and retrieval). \color{review}The broader impact spans greater than current tasks in \color{black}application (\eg kin-based generative modeling~\cite{gao2019will, ozkan2018kinshipgan}) and experimental settings (\eg with privacy a concern~\cite{mingaaai2020}). \color{review}\ac{rfiw} met the difficulty and practicality of today; \color{black}the question how best to formulate the problem remains an open research question. As such, this survey aims to provide a stronghold on the laboratory-style evaluations as seen appropriate in the modern day.

\subsection{Future work}\label{sec:next-steps}
It is an exciting and opportune time for kin-based problems for researchers and practitioners alike. For starters, there is a lot of room for improving \ac{sota}, and even the experiments (\ie design, purpose, and extent). For instance, incorporating additional label types (\ie other soft attributes like expression, age, and ethnicity), different data splits and protocols (\eg given a father, daughter, and grandparents from the side of the mother, determine the mother), and practical use-cases (\eg automate family photo-album creation). Generative-based tasks also hold promise in directions to take next: whether improved predictive capability of a child's face - provided a pair of parents, or a more fine-grained view of predicting any node in a family tree - provided samples of all other family members - then the room for improvement and potential for growth is furthered.


\noindent\textbf{Fairness.} A few recent attempts have been made by researchers to address fair AI and transparency in kinship understanding. For instance, the latest version of \ac{rfiw}, supplemental to this survey, \ac{fiw} is now supported with a \emph{datasheet}: ``datasheets for datasets''~\cite{gebru2018datasheets}.\footnote{{\href{https://web.northeastern.edu/smilelab//fiw/fiw_ds.pdf}{https://web.northeastern.edu/smilelab/fiw/fiw\_ds.pdf}}} The motivation of datasheets is to promote transparency and, thus, to minimize the doubt from unknown biases that come and are inherited by publicly available data resources.
Specifically, \emph{datasheets} completely spec-out the data (\eg motivation, composition, collection process, preprocessing, updates, legal and ethical considerations). There are other methods for transparency that have been recently proposed with a similar motivation as ``datasheets for datasets'', such as \emph{fact sheets}~\cite{arnold2019factsheets} and \emph{model cards}~\cite{DBLP:journals/corr/abs-1810-03993}. \diff{Per Hanley~\textit{et al}., ``\emph{Offensive associations can be latent in popular machine learning datasets.''}\cite{hanley2020ethical}: where they support the use of cards to formalize the construction of datasets, but argue to especially \emph{people-centric datasets} hone-in on specific components that make-up their ``ethical highlighter''.}

\noindent\textbf{Privacy.} 
\color{black}Yet another area deserving of attention is privacy. As is the case for many \ac{ml} tasks, privacy has motivated researchers. Recently, Kumar~\etal proposed using a \ac{gnn} to first achieve \ac{sota} in family classification, and to then add noise to encrypt the data, and demonstrating that a variant of the model safely encapsulates the learned knowledge (\ie an ability to accurately deceiver)~\cite{mingaaai2020}.

\noindent\textbf{Social and cultural.}
A near radical piece of its time, Goode~\cite{goode1963world} surveyed family structure as more of a complex system than the `conjugal family form' of many traditional cultures (\eg Western, Chinese, Arab). Hence, we currently look at sets of persons as being either related or not related-- this does not account for the realistic setting that would be faced in the modern world. For instance, tri-subject pairs are structured by using a parent pair and a child (\ie using evidence of both parents)-- a scenario that is certainly a step in the right direction, as parents are often inferred through records on marriage and offspring. However, families are dynamic in many modern cultures-- step siblings and parents are common. Considering the setting of tri-subject: what happens if the father is true, but the mother is not, or vice versa-- a concept that propagates to all levels of the problem, and especially when considering complete family trees with connections to in-laws. Thus, the problem remains: how to best weight (\ie fuse~\cite{zhang2019feature}) different relationship types. Even simple questions have soft, varying solutions ~\cite{BREDART1999129} like \emph{Do we look more like our father?} 

\noindent\textbf{Feature fusion.} Still today, the underlying question remains. \emph{How to best fuse prior knowledge?} For instance, in tri-subject verification, an active research question concerns the fusion of the features from the two parents. Flipping this very problem around (\ie given parents, generate the face of the child), the question of feature fusion is still prominent. Looking ahead at attempts to solve the fine-grained problem of populating family trees, regardless if viewed as discriminate or generative, the question remains: \emph{how to best leverage prior knowledge of additional family members relatively of different types and degrees?}

Although the number of methods is great-- whether metric-learning, deep features, a variant of both-- most recent attempts only differ in the broad sense. Bottom-line, successes in all tasks have been tributes of systems based on a Siamese network(s) that encodes inputs from image-to-feature space. The feature space learned typically differs in the point and method of fusion. Specifically, paired samples are usually split evenly (\ie the number of pair-types of type \emph{KIN} and \emph{NON-KIN} for each relationship type is split \emph{fifty-fifty}). Provided a Siamese network, often pre-trained on auxiliary face recognition dataset, act as face encoders. In order to transform from feature-to-score space, either a metric, fusion technique, or both are applied-- this tends to be where methods differ, yet the same conventional coarse system holds (\ie Fig.~\ref{fig:siamese}). In summary, it is the Siamese net to encode faces, followed by some means of feature-fusion that are stove-piped to a metric or learning objective. Hence, some relevant aspects of such a system produce current \ac{sota} from which we had drawn conclusions, and especially in identifying research trends and open issues. We consider the most relevant among aspects for achieving effective systems as follows: (1) effective method for fusion; (2) representation that considers the relationship's direction; (3) detecting other attributes (\eg age and gender) and knowledge of the higher-level scene (\eg face detected in picture with car styles that hint the picture was taken in the 1970s).

\color{review}
\noindent\textbf{Multimodal data.} Let us consider other signals that can define visual data; let us consider other label types for faces that could also enhance performance. For instance, expressions and mannerisms are often similar for parent and child (\eg \emph{they have the mother's smile}). More complex dynamics for individual expressions and mannerisms can be effectively captured in video data~\cite{kollias2020analysing}. Hence, added knowledge that complements the visual information has proven useful for boosting kin-based recognition ratings~(\eg 3D facial images\cite{vijayan2011twins}, voice~\cite{wu2019audio}, MM~\cite{robinson2020familiesinMM}).\color{black}

\color{review}\noindent\textbf{Family synthesis.}The existing technology for synthesizing family members is still immature and generalization remains unsolved. A system should accept one-to-many members of a family tree and synthesize the appearances of the desired relative type. Still, many questions remain: how to fuse, handle dynamic inputs, the optimal way to reason about a family tree, and more.\color{black}


\subsection{Conclusion}
\color{review}
Surveying a decade of research in visual kinship recognition showed increasing interest with an increase in data resources. Clearly, the problems alone are challenging, even when compared to other machine-vision tasks (\eg conventional \ac{fr}). Furthermore, the task of designing, collecting, and annotating labels is exceptionally difficult for kin-based problems. Thus, as contributions in data are proposed, interest seems to spike in response. With the release of the large-scale \ac{fiw} dataset, for the first time, a data resource attempts to closely mimic data distributions of families around the globe. Moreover, it meets the capacity of the modern day, deep learning models. \ac{fiw}, having had many existing datasets to learn from, remains the largest and most dynamic. However, \color{black}the release of \ac{fiw} was only the beginning, as efforts were then spent on annual challenges (\ie four consecutive years, 2017-2020, and also a Kaggle competition). With the resource and incentive provided by challenges, motivation for researchers to engage is ever so high and thus, we present this survey-  not only as a means to realize the aspects that have been effective and vice versa-  but also to provide a solid foundation for the next decade to build upon well-defined protocols and problem statements, each supported with source code in a single location, enabling even a wider audience to get started and contributing to the problem.

The deep learning revolution has only begun for visual kinship recognition - how to embed, how to fuse, how to interrupt - how do experts across disciplines engage by leveraging for a deeper understanding in inheritance from a strictly scientific point-of-view (\ie anthropology)! Hence, if we can devise the right tools for the right scholars synergy is bound to reveal insights in the nature of faces within families. Considering the many benchmarks that have a lot of room for improvement, along with the many social and relational data mining that is made possible with soft-attribute labels such as those in \ac{fiw}, it is an exciting time for junior, senior, and practical researchers to reap benefits alongside its place with pure business, product, and patent design.

We covered (1) progress in automatic kinship understanding and the major milestones, (2) problem statements as a foundation for consistent and fair comparisons moving forward, (3) \ac{sota} and top scores, (4) technical challenges, (5) practical value of various tasks, (6) forecasts for research in the foreseeable future. We encourage readers to see the open-source project to use, reference, and contribute. We challenge the experts and invite newcomers to join the effort, and to consider the next steps highlighted. Let us keep moving forward on this unique biometric problem.

